\documentclass[11pt]{article}

\PassOptionsToPackage{table}{xcolor}
\usepackage[final]{acl}

\usepackage{times}
\usepackage{latexsym}
\usepackage{enumitem}

\usepackage{amsfonts}
\usepackage[ruled,vlined,linesnumbered]{algorithm2e}
\usepackage{booktabs}
\usepackage{amsmath}
\usepackage{multirow}
\usepackage[most]{tcolorbox}
\usepackage{makecell}
\usepackage{needspace}  
\usepackage{dblfloatfix}   
\usepackage[section]{placeins} 
\usepackage{float}   
\usepackage{adjustbox}

\definecolor{RoyalBlue}{RGB}{65,105,225}
\definecolor{ForestGreen}{RGB}{34,139,34}
\definecolor{Orange}{RGB}{255,165,0}

\tcbset{
  enhanced,
  breakable,
  boxrule=0.6pt,
  arc=1.2mm,
  left=1.2mm,right=1.2mm,top=0.9mm,bottom=0.9mm,
  before skip=6pt, after skip=8pt,
  fontupper=\normalsize\normalfont
}

\newtcolorbox{AppBox}[2][]{colback=black!2,colframe=black!55,fonttitle=\bfseries,fontupper=\normalfont,title={#2},#1}
\newtcolorbox{PromptBox}[2][]{colback=RoyalBlue!3,colframe=RoyalBlue!55,fonttitle=\bfseries,fontupper=\normalfont,title={#2},#1}
\newtcolorbox{FormatBox}[2][]{colback=Orange!4,colframe=Orange!60,fonttitle=\bfseries,fontupper=\normalfont,title={#2},#1}
\newtcolorbox{CaseBox}[2][]{colback=ForestGreen!3,colframe=ForestGreen!55,fonttitle=\bfseries,fontupper=\normalfont,title={#2},#1}

\definecolor{BRIcol}{RGB}{201,217,240} 
\definecolor{NPScol}{RGB}{204,235,218} 
\definecolor{RVcol}{RGB}{245,239,188}  
\definecolor{Polcol}{RGB}{246,216,172} 
\definecolor{CRScol}{RGB}{217,183,223} 
\definecolor{SoVcol}{RGB}{238,179,200} 

\usepackage[T1]{fontenc}
\usepackage[utf8]{inputenc}
\usepackage{microtype}
\usepackage{inconsolata}
\usepackage{graphicx}
\usepackage{url}
\usepackage{textcomp}

\title{BrandEval: A Two-Track Multi-Agent Benchmark for Risk-Sensitive \\ LLM Crisis Communication}

\author{
  \textbf{Haotian Gan}\textsuperscript{1\textsuperscript{*}},
  \textbf{Weidong Tang}\textsuperscript{2\textsuperscript{*}},
  \textbf{Hanbin Sun}\textsuperscript{2},
  \textbf{Yi Luo}\textsuperscript{3},
  \\
  \textbf{Xintian Li}\textsuperscript{4},
  \textbf{Jiayong Wang}\textsuperscript{5},
  \textbf{Wanyue Li}\textsuperscript{6},
  \textbf{Yudong Li}\textsuperscript{4\textdagger}
\\
\\
  \textsuperscript{1}Shaanxi University of Science and Technology,
  \textsuperscript{2}China Agricultural University,
\\
  \textsuperscript{3}Tianjin University,
  \textsuperscript{4}Tsinghua University,
\\
  \textsuperscript{5}Shaanxi Normal University,
  \textsuperscript{6}Hong Kong Metropolitan University
\\
  \textsuperscript{*}Equal Contribution.
  \quad
  \textsuperscript{\textdagger}Corresponding author.
}

\usepackage{titlesec}

\titlespacing*{\section}
  {0pt}
  {1.15ex plus 0.35ex minus 0.2ex}
  {0.70ex plus 0.2ex}

\titlespacing*{\subsection}
  {0pt}
  {0.90ex plus 0.25ex minus 0.15ex}
  {0.45ex plus 0.15ex}

\titlespacing*{\subsubsection}
  {0pt}
  {0.70ex plus 0.2ex minus 0.1ex}
  {0.35ex plus 0.1ex}

\titlespacing*{\paragraph}
  {0pt}
  {0.55ex plus 0.15ex minus 0.1ex}
  {0.75em}

\begin{document}
\maketitle
\begin{abstract}
Evaluating LLMs for crisis communication requires more than judging isolated responses, because public feedback and competitor actions can reshape trust over time. Existing benchmarks remain largely static and single-turn, offering limited evidence about long-horizon risk-sensitive communication. We introduce \textbf{BrandEval}, a two-track benchmark for product-crisis communication, focusing on public brand responses to product-related safety, quality, service, and trust incidents. BrandEval-Static evaluates stakeholder-aware strategic analysis, while BrandEval-Dynamic stress-tests response policies in \textbf{BrandPolis}, a controlled multi-agent sandbox with consumer feedback, competitor responses, and trust-risk trajectories. We also construct \textbf{BrandSRD}, a Chinese crisis-response dataset, and instantiate a reference Strategic Agent using \textbf{Strategic Rationale (SR)}. Across proprietary and open-source models, we find that static strategic quality is informative but incomplete: it aligns with some end-state outcomes but does not reliably predict volatility and delayed recovery; dynamic stress tests further expose tail-risk failures under manipulative or weakly grounded policies. The code and dataset will be available at https://github.com/htgan-ai/BrandEval.
\end{abstract}

\section{Introduction}
\begin{figure}[t]
\centering
\includegraphics[width=0.46\textwidth]{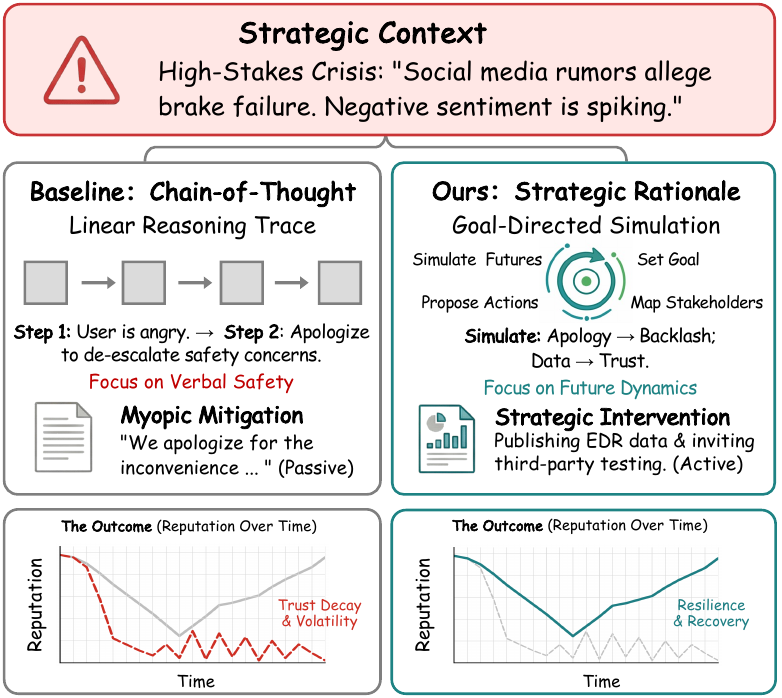}
\caption{\textbf{Divergent Futures.} Under the same crisis context, different response formulations can lead to divergent long-horizon trust trajectories in simulation.}
\label{fig:teaser}
\end{figure}

When a product crisis happens, such as safety issues or service failures, a company's public response can strongly affect how people judge its responsibility, safety, and trustworthiness, especially when the incident unfolds across social media, news, and official channels~\cite{coombs2007protecting, leskovec2007dynamics}.
This is especially important on social media, where public opinion can change quickly and spread widely. A slow or vague response may increase public anger~\cite{coombs2007ongoing}. 
In contrast, a clear and responsible response may help reduce uncertainty and support trust recovery~\cite{hallahan2007defining}. 
For this reason, crisis communication should not be treated as a simple text-generation task. 
The key question is not only whether a response sounds good, but whether it can help the company handle the crisis over time~\cite{entman1993framing}.
Figure~\ref{fig:teaser} illustrates why crisis communication requires evaluating response consequences over time, not isolated message quality alone.

Large language models (LLMs) are increasingly used to support business communication, public relations, and risk management~\cite{li2024schema}. 
Although Chain-of-Thought (CoT) methods can make strategic reasoning more inspectable~\cite{wei2022chain, jaech2024openai, guo2025deepseek}, current LLM evaluations mostly assess a single answer in a static setting and are therefore not well suited for crisis communication~\cite{zhang2024llm}.  
They can measure whether a model gives a reasonable analysis or writes a fluent response, but they lack a dynamic framework for evaluating how communication strategies perform under public feedback and competitive pressure~\cite{grewal2025brands}. 
This motivates our main question: \emph{How can we evaluate whether LLMs can make credible, responsible, and effective crisis-communication decisions over time?}

To study this question, we introduce \textbf{BrandEval}, a two-track benchmark for risk-sensitive product crisis communication. 
The first track, \textbf{BrandEval-Static}, evaluates single-turn strategic analysis. 
It tests whether a model can understand the crisis context, identify key stakeholders~\cite{pendharkar2012game}, reason about risks, and propose practical actions. 
The second track, \textbf{BrandEval-Dynamic}, stress-tests response policies in \textbf{BrandPolis}, a controlled multi-agent environment.
In BrandPolis, a focal brand interacts with consumer agents, competitor agents, and a platform environment over multiple rounds. 
This setup allows us to observe how different communication strategies affect trust, risk, reputation stability, and recovery over time~\cite{ji2025systematic}.
We also construct \textbf{BrandSRD}, a crisis-response dataset with de-identified decision points, real and counterfactual response options, preference labels, and explanatory rationales. 
Based on this dataset, we build a reference Strategic Agent guided by \textbf{Strategic Rationale}, which encourages goal setting, stakeholder tracking, and future-aware action selection.

Our experiments show that BrandEval-Static can diagnose whether a model produces a plausible crisis analysis, but it cannot measure whether the proposed actions remain effective after public reactions and competitor responses unfold. 
BrandEval-Dynamic complements this by revealing long-term trust trajectories, reputation volatility, recovery speed, polarization, and tail-risk failures. 
These results show that crisis communication agents need more than fluent writing or one-shot reasoning. 
They need to make credible, evidence-based, and socially responsible decisions across time.

The main contributions are as follows:
\begin{itemize}
    \item We introduce \textbf{BrandEval}, a two-track benchmark pairing static stakeholder diagnosis with dynamic multi-agent evaluation for risk-sensitive product crisis communication.
    \item We develop \textbf{BrandPolis} and \textbf{BrandSRD} to support long-horizon simulation, controlled scenario construction, and preference-grounded evaluation.
    \item We show that static and dynamic evaluations are complementary: dynamic stress testing captures trajectory stability and tail risks that static diagnostic scores cannot assess.
\end{itemize}

\section{Related Work}
\begin{figure*}[t]
\centering
\includegraphics[width=0.96\textwidth]{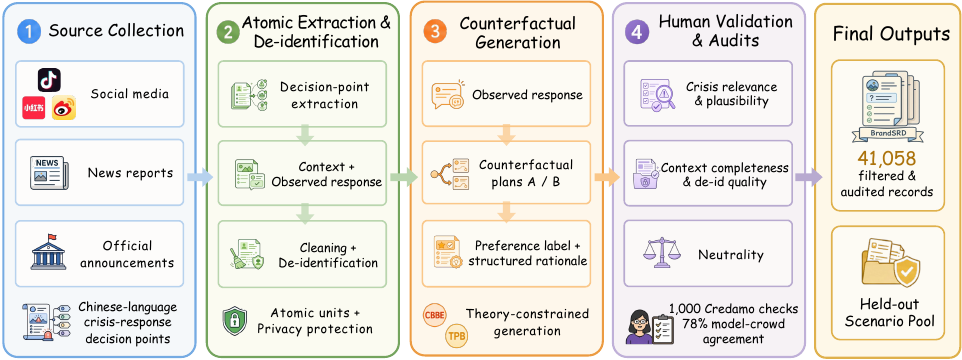}
\caption{\textbf{BrandSRD construction pipeline.} Source collection is followed by atomic extraction and de-identification, counterfactual generation, and human validation and audits, yielding 41,058 filtered and audited records and a held-out evaluation pool for BrandEval.}
\label{fig:brandsrd-pipeline}
\end{figure*}

Recent work on LLM agents with planning and social reasoning has used language models to simulate interactive social systems~\cite{hao2023reasoning, kosinski2023theory}.
This simulation paradigm has been extended to commercial contexts, including business decision making~\cite{li2025metaagents, hazenberg2025multi}, commercial communication and market interaction~\cite{karande2024persuasion}, and quantifying perceptual gaps between brands and user-generated content~\cite{gan2025aligned}.
Social simulation systems further model realistic social network dynamics with multimodal agents~\cite{touzel2025sandboxsocial} and scale up large social simulations~\cite{tang2025gensim}.
These studies provide useful interaction substrates, but rarely stress-test accountable crisis communication with competitive responses.

Agent evaluation has also moved beyond static answer checking, as static QA benchmarks often miss interaction competence and remain vulnerable to contamination~\cite{xu2024benchmark}.
Recent work addresses this issue through procedural instance generation~\cite{fan2023nphardeval, kurtic2024mathador} and benchmarks built from time-varying real-world data, such as historical knowledge snapshots and user dialogues~\cite{ouyang2025hoh, bayat2024factbench}.
Agent evaluation has expanded from general multi-agent suites such as MultiAgentBench~\cite{zhu2025multiagentbench} to dynamic settings, including asynchronous environments where events evolve independently of agent actions~\cite{froger2026gaia} and simulated commercial marketplaces with longitudinal competitive dynamics~\cite{kim2026evaluation}.
Yet these benchmarks do not directly assess public-facing crisis response or the trust and risk trajectories shaped by communication policies.

Domain-focused agent benchmarks have begun to cover enterprise workflows~\cite{vishwakarma2025can, gupta2026world}, customer support~\cite{balaji2026beyond}, and advertising analytics~\cite{hu2026ad}.
Crisis informatics and operational communication, in turn, evaluate practical deliverables from social streams, including situation assessment and information triage~\cite{alam2021crisisbench}, crisis timeline extraction and summarization~\cite{rajaby2022crisisltlsum}, and online public opinion report generation~\cite{yu2025opor}.
Together, these lines cover business task execution and crisis information processing, but not how public-facing responses reshape stakeholder reactions, competition, and recovery.
Motivated by this gap, BrandEval contributes an evaluation protocol rather than a social simulator alone: it pairs rubric-based static diagnosis with controlled dynamic stress tests of accountability, verifiability, de-escalation, and long-horizon process health.

\section{BrandSRD Dataset}
\subsection{BrandSRD Construction}
\paragraph{Dataset overview.}
BrandSRD is a Chinese-language dataset of 41{,}058 crisis-response decision points constructed from public crisis-related posts and event context; Table~\ref{tab:data-stats} reports the source distribution. It is derived from user posts on three Chinese social-media platforms (Weibo, Douyin, and Xiaohongshu) collected from 2023-12-15 to 2025-12-15, and supplemented with event context from news reports and official announcements.
Figure~\ref{fig:brandsrd-pipeline} summarizes the BrandSRD construction pipeline and a separate held-out evaluation pool reserved from the same source collection for later benchmark evaluation.
Each data record includes a context, one observed or de-identified brand response, two counterfactual response plans, a preference label, and a rationale.

\begin{table}[b]
\centering
\caption{\textbf{BrandSRD source distribution.} Counts of posts used to construct BrandSRD across 12 product domains and three platforms (XHS = Xiaohongshu).}
\small
\setlength{\tabcolsep}{5pt}
\begin{tabular}{lcccc}
\toprule
\textbf{Product} & \textbf{Weibo} & \textbf{Douyin} & \textbf{XHS} & \textbf{Total} \\
\midrule
Smartwatch     & 1078 &  786 &  482 & 2346 \\
Refrigerator   & 1024 & 1125 &  741 & 2890 \\
Air conditioner & 1096 &  553 &  781 & 2430 \\
Rice cooker    &  858 & 2517 &  676 & 4051 \\
Smart door lock& 1360 & 2268 &  449 & 4077 \\
Translation pen & 1265 & 1051 &  614 & 2930 \\
Smartphone     & 1177 &  734 & 1847 & 3758 \\
Television     & 1055 &  929 &  921 & 2905 \\
Speaker        & 1606 &  563 & 1343 & 3512 \\
Dishwasher     & 1721 &  580 &  465 & 2766 \\
Personal computer       &  931 &  948 &  872 & 2751 \\
Electric vehicle    & 5548 &  636 &  458 & 6642 \\
\bottomrule
\end{tabular}
\label{tab:data-stats}
\end{table}

\paragraph{Scenario construction.}
Each record is constructed through a controlled scenario-construction pipeline. The pipeline first filters posts and event fragments for crisis-response decision points, where brand communication choices may affect stakeholder trust, public reaction, and recovery. For each retained decision point, we use Gemini-2.5-Pro to extract the crisis context and observed brand response, then de-identify and neutrally rewrite the case into a self-contained scenario while preserving the commercial context, stakeholder trade-offs, and underlying claims. Based on this scenario, Gemini-2.5-Pro generates two counterfactual response plans under Customer-Based Brand Equity (CBBE)~\cite{keller1993conceptualizing} and Theory of Planned Behavior (TPB)~\cite{ajzen1991theory} constraints. Each record therefore contains one observed or de-identified response and two plausible counterfactual plans for the same scenario; the preference label selects the preferred option among these candidates and is accompanied by a structured rationale.

\begin{figure*}[htbp]
\centering
\includegraphics[width=0.96\textwidth]{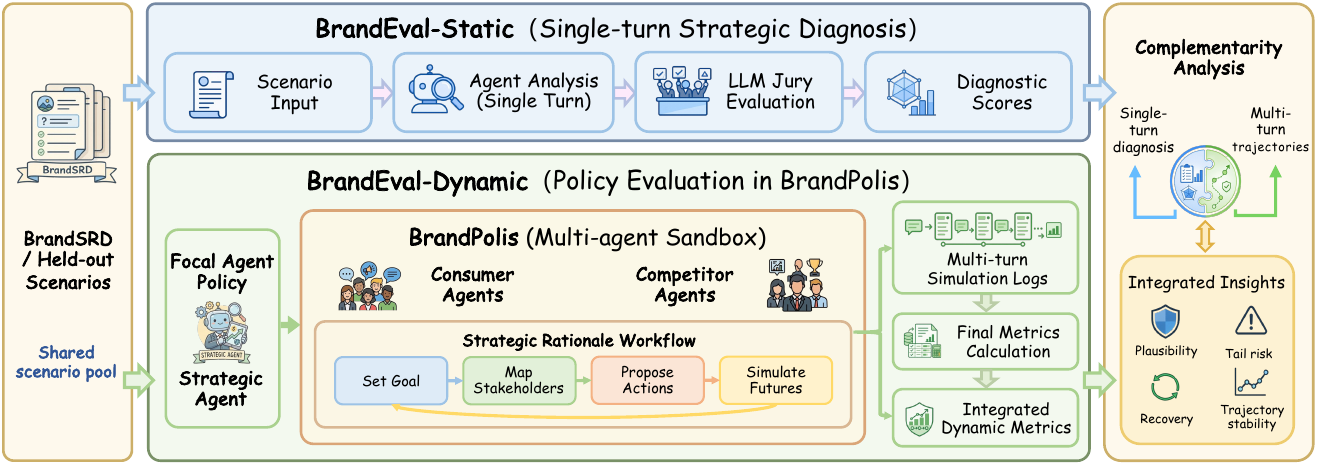}
\caption{\textbf{Overview of BrandEval.} BrandEval pairs Static diagnosis on held-out scenarios with Dynamic policy evaluation in BrandPolis. Strategic Rationale guides the Strategic Agent, which serves as a reference policy within the Dynamic track.}
\label{fig:overview}
\end{figure*}

\subsection{Data Quality and Evaluation Audits}
\label{sec:data_quality_audits}
To assess the plausibility and quality of BrandSRD and the held-out evaluation pool, we audit scenario screening and preference labels. For scenario screening, generated scenario briefs are reviewed for crisis relevance, commercial plausibility, contextual completeness, de-identification quality, and neutrality. Three domain reviewers independently audit 600 randomly sampled scenario briefs, achieving 93.3\% agreement and Fleiss' $\kappa = 0.84$. Remaining disagreements are adjudicated by an additional senior reviewer. Rejected cases mainly involve implausible business settings or market behavior (35\%), ambiguous links between crisis events and stakeholder reactions (30\%), insufficient context (20\%), and de-identification or safety issues (15\%).

For preference labels, we validate 1{,}000 randomly sampled BrandSRD records through three-way structured crowd questionnaires on Credamo,\footnote{\url{https://www.credamo.com}} obtaining a 78\% model--crowd agreement rate, a mean winning vote share of 0.633, and a mean top-two vote share gap of 0.433. These audits assess scenario plausibility, construction quality, and inter-rater agreement. Additional diagnostics are provided in Appendix.

\section{BrandEval Benchmark}
This section presents how BrandEval operationalizes crisis communication evaluation through two complementary tracks: BrandEval-Static evaluates single-turn strategic analysis, BrandEval-Dynamic evaluates response policies over multiple rounds, and BrandPolis provides the controlled environment for dynamic evaluation. Figure~\ref{fig:overview} gives an overview of BrandEval.

\subsection{BrandEval-Static}
\label{sec:brandeval-static}
\begin{figure}[b]
\centering
\includegraphics[width=0.48\textwidth]{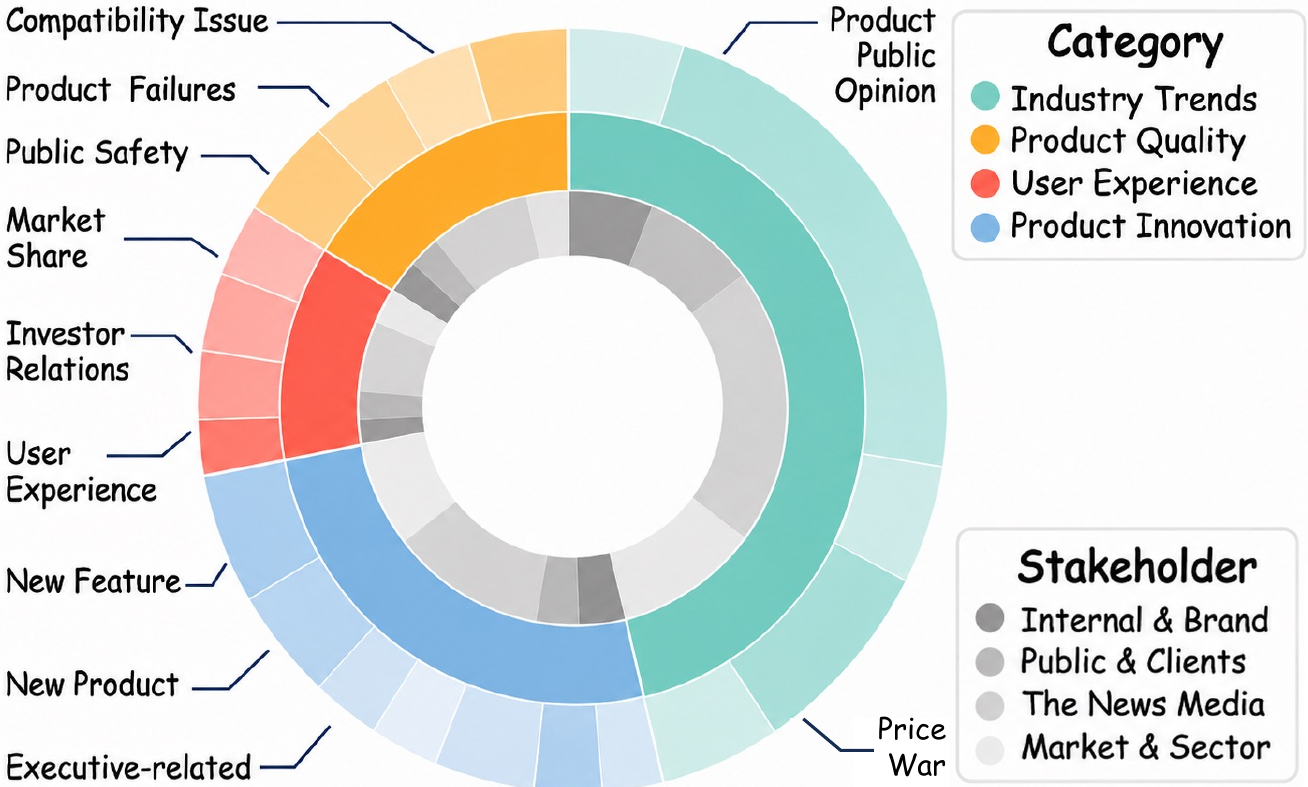} 
\caption{\textbf{BrandEval prototype distribution.} Topic and stakeholder composition of 1{,}000 scenarios.}
\label{fig:benchmark}
\end{figure}

BrandEval-Static evaluates single-turn strategic diagnosis on 1{,}000 self-contained scenarios drawn from the held-out evaluation pool. Figure~\ref{fig:benchmark} summarizes the underlying prototype pool's topic and stakeholder composition.

\paragraph{Evaluation Setup.}
For each scenario, the model under test acts as a strategic consultant. It receives a self-contained scenario description and outputs a strategic analysis report. The report must diagnose stakeholder impacts across consumers, the focal brand, public and media actors, and competitors, then synthesize these trade-offs into an actionable recommendation. This setup tests whether the model can identify affected stakeholders, use evidence responsibly, anticipate competitive dynamics, and formulate a coherent response plan.

\paragraph{Scoring Protocol.}
We score each report with a stakeholder-theory-inspired rubric~\cite{freeman2010strategic} covering four dimensions: Consumer \& Brand, Public \& Media, Competitor, and Strategic Synthesis. Given the open-ended nature of strategic reasoning, exhaustive expert scoring is costly to scale. We therefore use a three-LLM judge panel following LLM-as-judge practice~\cite{liu2023g}. GPT-5-mini, DeepSeek-v3.1, and Claude-4.5-Haiku independently score each report on the four dimensions using a 1--10 scale, and we average their scores per scenario and dimension. Against human scores on 300 sampled strategic-analysis outputs, the panel obtains a mean absolute error of 1.08 points and a Pearson correlation of 0.64. An evidence-sensitive audit further separates verifiable strategies from specious ones with Cohen's $d = 1.34$ and an 89\% win rate.

\subsection{BrandEval-Dynamic}
\label{sec:brandeval-dynamic}

\paragraph{Evaluation Setup.}
BrandEval-Dynamic evaluates crisis-response policies through 20-round interactive episodes in BrandPolis. We select 50 scenarios from the same held-out evaluation pool via stratified sampling over topics and risk profiles. For each scenario, the evaluated model serves as the focal crisis-response agent and interacts with consumer agents, competitor agents, and the platform environment. Each published action can affect subsequent stakeholder feedback, competitor responses, and platform-visible market briefs. We use a fixed horizon $T=20$ and run each model-scenario pair with $K=5$ random seeds, reporting seed-averaged metrics.

\paragraph{Reference Strategic Agent.}
We instantiate a reference Strategic Agent for BrandEval-Dynamic using Strategic Rationale, a structured workflow for round-by-round crisis response.
Each round, the agent first states a verifiable goal that balances business needs with social responsibility; this goal then constrains stakeholder tracking, candidate planning, and action selection, encouraging commitments that remain credible under public scrutiny~\cite{carroll1991pyramid}. The agent then summarizes the stakeholder state, drafts publishable response candidates, and selects the option that best balances long-horizon credibility and risk for publication.

\subsection{BrandPolis Environment}
\label{sec:brandpolis}
BrandPolis implements BrandEval-Dynamic as a controlled stress-test environment for relative policy evaluation. It is a discrete-time multi-agent sandbox in which published brand actions, consumer feedback, competitor responses, and platform-mediated exposure jointly shape comparable trust and risk trajectories.

\paragraph{Consumer Agents.}
The population consists of consumer agents whose static personas follow a $5 \times 3$ factorial design over decision drivers and brand stances: five primary decision drivers (performance, value for money, user experience, social identity, and safety)~\cite{engel1986consumer}, crossed with three initial stance levels (supporters, neutrals, detractors)~\cite{gera2025deep} that may shift as new crisis evidence updates their advocacy intent, in line with cognitive dissonance effects~\cite{aronson1969theory}. Each agent maintains a five-dimensional internal state that combines three perception dimensions from CBBE (competence belief, value congruence, emotional connection) with two intention dimensions from TPB (purchase intent, advocacy intent). Consumer states evolve via a first-order exponential moving average driven by message-derived impacts:
\begin{equation}
\label{eq:ema_main}
\mathbf{V}_{u}^{(t)} = (1-\lambda_u)\mathbf{V}_{u}^{(t-1)} + \lambda_u \gamma^{(t)}  \tanh\!\big( \mathbf{X}_{u}^{(t)}\big)
\end{equation}
where $\lambda_u\in(0,1)$ controls update responsiveness, $\mathbf{X}_{u}^{(t)}$ is an auditable text-to-state impact signal estimated from message features, and $\gamma^{(t)}\in[0,1]$ is an evidence-consistency gate for official claims. We treat $\mathbf{V}_{u}^{(t)}$ as simulation proxies for belief and intention.

\paragraph{Focal and Competitor Agents.}
The focal agent is the policy under evaluation: at each round, it receives the current market brief and produces a publishable crisis-response action for the focal brand. This policy can be a zero-shot LLM, a prompted agent, or the SR-guided Strategic Agent. BrandPolis also includes several competitor agents, all operating under bounded rationality~\cite{simon1955behavioral}.
Competitor agents follow four archetypes, Prospector, Defender, Analyzer, and Reactor, which adapt Miles and Snow's strategic typology~\cite{miles1978organizational} into utility-driven personas with distinct goals and risk preferences. For each scenario, we specify which competitor brands are present and assign one archetype to each. The focal agent and all competitor agents alternate between external observation, where they track topics and platform-visible reputation proxies, and internal monitoring of their own published commitments. Evaluation tracks how actions shift belief and intent trajectories, while the next-round prompt uses the market brief to convert current market intelligence into a concrete strategic action.

\paragraph{Platform Environment.}
The platform runs in discrete rounds and mediates exposure under attention scarcity. At round $t$, it collects brand, competitor, and user-generated messages. Official brand posts are treated as broadcast and prioritized in consumers' feeds, while other items are sampled by sender influence and topic heat, mimicking weak-tie diffusion in crises~\cite{granovetter1973strength}. Each consumer has feed capacity $C$ and processes at most $C$ items, with processing probability modulated by susceptibility and stance alignment; processed messages update CBBE and TPB internal states via Eq.~\eqref{eq:ema_main}. The platform produces a market brief of platform-visible aggregates and dominant topics as the next observation $S_{t+1}$; it is built only from the public message log and engagement counters and excludes latent belief states and evaluation metrics. To discourage unsupported evidence claims, BrandPolis applies an evidence-consistency gate $\gamma^{(t)}\in[0,1]$ that attenuates the impact of unsupported evidence claims in official actions.

\paragraph{Evaluation Metrics.}
We evaluate dynamic performance with six metrics grouped into end-state outcomes and trajectory-level process-health signals. As public-interest outcomes, Brand Reputation Index (BRI)~\cite{fombrun2000reputation} operationalizes population-level trust as an aggregated reputation signal over CBBE-aligned perception dimensions. Net Promoter Score (NPS)~\cite{reichheld2003one} operationalizes advocacy as a population-level summary of the TPB-aligned intent signal. Share of Voice (SoV)~\cite{farris2010marketing} contextualizes competitive attention as the brand's share of crisis-topic-related discussion volume within the sandbox. SoV is an attention measure and has no uniform preferred direction. As process-health signals, Polarization Index (Pol.)~\cite{esteban1994measurement} measures stance polarization across supporters, neutrals, and opponents. Average Reputation Volatility (ARV)~\cite{leberknight2011volatility} captures volatility of the reputation trajectory over the full horizon, and Crisis Recovery Speed (CRS)~\cite{coombs2007protecting} measures how quickly reputation recovers after reaching its episode minimum. We report per-round trajectories and summarize them with end-of-horizon values ($BRI_T$, $NPS_T$, $SoV_T$, $Pol_T$) and trajectory-level summaries (ARV, CRS), enabling comparisons between final outcomes and process health.

\section{Experiments}
\subsection{Experimental Setup}
\label{sec:exp_setup}
We evaluate mainstream proprietary and open-source LLMs under the same BrandEval protocol, using zero-shot prompting for direct LLM baselines. BrandEval-Static follows the stakeholder-aware scoring protocol in Section~\ref{sec:brandeval-static}, and BrandEval-Dynamic follows the interaction protocol in Section~\ref{sec:brandeval-dynamic}. In addition to zero-shot baselines, we include a reference Strategic Agent initialized from Qwen3-8B~\cite{yang2025qwen3}, an open-weight backbone well suited to fine-tuning in BrandSRD's Chinese setting. It is trained only on the BrandSRD training split and evaluated on disjoint BrandEval held-out scenarios with the Strategic Rationale workflow; training details are provided in Appendix.

\subsection{BrandEval-Static Results}
We first report results on BrandEval-Static, which evaluates single-turn strategic analysis quality under a multi-LLM judge panel (Table~\ref{tab:static-main}). Two findings are salient. First, frontier general-purpose models dominate static scores, consistent with the Static track rewarding immediate diagnostic coverage and rhetorically complete reports in a single turn. Second, the Strategic Agent is competitive with strong open models but does not surpass the best proprietary systems; this gap is informative rather than surprising, because SR is optimized for iterative crisis handling rather than maximizing one-shot narrative completeness. 

\begin{table}[h]
\centering
\caption{\textbf{BrandEval-Static results.} Multi-LLM panel scores (1--10) averaged over all scenarios. C/B = Consumer \& Brand; P/M = Public \& Media; Comp. = Competitor; Strat. = Strategic Synthesis.}
\small
\setlength{\tabcolsep}{6pt}
\begin{tabular}{lcccc}
\toprule
\textbf{Model} &
\textbf{C/B} &
\textbf{P/M} &
\textbf{Comp.} &
\textbf{Strat.} \\
\midrule
GPT\text{-}5.1                   & 7.86   & 7.81   & 7.42   & 7.60   \\
Gemini\text{-}2.5\text{-}Pro     & 7.61   & 7.64   & 7.28   & 7.34   \\
Claude\text{-}Sonnet\text{-}4.5  & 7.52   & 7.71   & 7.36   & 7.40   \\
\midrule\midrule
GLM\text{-}4.5                   & 7.11   & 7.16   & 6.93   & 6.89   \\
DeepSeek\text{-}R1               & 6.74   & 6.80   & 6.46   & 6.47   \\
R1\text{-}Distill\text{-}Qwen\text{-}7B
                                 & 4.88   & 4.31   & 4.44   & 4.04   \\
Qwen3\text{-}8B                  & 6.20   & 6.12   & 6.00   & 5.73   \\
Qwen3\text{-}14B                 & 6.67   & 6.59   & 6.61   & 6.44   \\
Llama\text{ }3.1\text{-}8B       & 4.39   & 3.90   & 3.89   & 3.52   \\
\midrule\midrule
\rowcolor{gray!15}
Strategic Agent                  & 6.75 & 6.73 & 6.51 & 6.44 \\
\bottomrule
\end{tabular}
\label{tab:static-main}
\end{table}

\subsection{BrandEval-Dynamic Results}

\begin{table*}[t]
\centering
\caption{\textbf{BrandEval-Dynamic results.} Mean over 50 scenarios under Single (Prospector) and Dual (Prospector+Analyzer) competitor regimes. For directed metrics, best values are \textbf{bold} and second-best values are \underline{underlined}; SoV is reported descriptively. $\uparrow$ indicates higher is better, while $\downarrow$ indicates lower is better.}
\small
\setlength{\tabcolsep}{3.5pt}
\begin{adjustbox}{width=\textwidth}
\begin{tabular}{l|cc|cc|cc|cc|cc|cc}
\toprule
\multirow{2}{*}{\textbf{Model}} &
\multicolumn{2}{c|}{\textbf{BRI$\uparrow$}} &
\multicolumn{2}{c|}{\textbf{NPS$\uparrow$}} &
\multicolumn{2}{c|}{\textbf{ARV$\downarrow$}} &
\multicolumn{2}{c|}{\textbf{Pol.$\downarrow$}} &
\multicolumn{2}{c|}{\textbf{CRS$\downarrow$}} &
\multicolumn{2}{c}{\textbf{SoV}} \\
\cline{2-13}
& \raisebox{-0.6ex}{\textbf{Single}} & \raisebox{-0.6ex}{\textbf{Dual}}
& \raisebox{-0.6ex}{\textbf{Single}} & \raisebox{-0.6ex}{\textbf{Dual}}
& \raisebox{-0.6ex}{\textbf{Single}} & \raisebox{-0.6ex}{\textbf{Dual}}
& \raisebox{-0.6ex}{\textbf{Single}} & \raisebox{-0.6ex}{\textbf{Dual}}
& \raisebox{-0.6ex}{\textbf{Single}} & \raisebox{-0.6ex}{\textbf{Dual}}
& \raisebox{-0.6ex}{\textbf{Single}} & \raisebox{-0.6ex}{\textbf{Dual}} \\
\midrule

\textit{Initial Value}
& -12.07 & -12.07
& -32.00 & -32.00
& - & -
& 82.41 & 82.41
& - & -
& 64.25 & 64.23 \\

\midrule\midrule

GPT\text{-}5.1
& \textbf{-2.34} & -3.62
& \textbf{-1.00} & \textbf{-2.00}
& 33.09 & 31.46
& \underline{1.98} & \textbf{0.00}
& \textbf{14} & \textbf{13}
& 68.27 & 68.94 \\

Gemini\text{-}2.5\text{-}Pro
& -2.89 & -3.16
& -3.00 & \textbf{-2.00}
& 34.29 & 33.97
& 3.92 & 7.68
& \underline{15} & \underline{14}
& 68.93 & 69.61 \\

Claude\text{-}Sonnet\text{-}4.5
& -3.03 & -3.75
& \underline{-2.00} & -4.00
& 32.36 & 31.11
& \underline{1.98} & \underline{1.98}
& 16 & 16
& 68.91 & 69.59 \\

\midrule\midrule

GLM\text{-}4.5
& -4.09 & \underline{-2.95}
& -3.00 & \underline{-3.00}
& 30.56 & 30.22
& 3.92 & \underline{1.98}
& \textbf{14} & 15
& 69.62 & 68.94 \\

DeepSeek\text{-}R1
& -4.04 & -5.28
& -6.00 & -4.00
& 28.93 & 28.06
& 5.82 & 3.92
& 16 & 16
& 67.65 & 75.96 \\

R1\text{-}Distill\text{-}Qwen\text{-}7B
& -4.26 & -3.42
& -9.00 & -6.00
& 27.80 & 30.72
& \underline{1.98} & 3.92
& 18 & 17
& 69.60 & 69.61 \\

Qwen3\text{-}8B
& -3.54 & -4.30
& -3.00 & -9.00
& 30.06 & 29.74
& \underline{1.98} & \underline{1.98}
& 16 & 15
& 63.46 & 68.90 \\

Qwen3\text{-}14B
& -4.01 & -5.10
& -5.00 & -7.00
& 29.01 & \underline{27.82}
& 3.92 & \underline{1.98}
& 17 & 16
& 68.89 & 71.84 \\


Llama\text{ }3.1\text{-}8B
& -5.11 & -5.43
& -4.00 & -8.00
& \underline{28.68} & \textbf{24.81}
& \textbf{0.00} & 5.82
& 18 & 20
& 69.90 & 65.05 \\

\midrule\midrule

\rowcolor{gray!15}
Strategic Agent
& \underline{-2.36} & \textbf{-2.76}
& \textbf{-1.00} & \textbf{-2.00}
& \textbf{27.78} & 31.83
& \underline{1.98} & \underline{1.98}
& \textbf{14} & \textbf{13}
& 69.61 & 69.99 \\

\bottomrule
\end{tabular}
\label{tab:dynamic-main}
\end{adjustbox}
\end{table*}

Building on the static diagnostic, we evaluate models in BrandEval-Dynamic to test whether single-turn analysis leads to robust long-horizon behavior under interaction (Table~\ref{tab:dynamic-main}). Three findings stand out. (i) \textbf{Trust gains require process health:} higher reputation and advocacy are more meaningful when accompanied by lower volatility, lower polarization, and faster recovery. For example, the Strategic Agent nearly matches GPT-5.1 on single-competitor trust ($BRI_T=-2.36$ vs. $-2.34$) while exhibiting lower volatility (ARV 27.78 vs. 33.09), indicating a more stable recovery path. (ii) \textbf{Dynamic metrics reveal trade-offs:} the Strategic Agent achieves competitive end-state trust and recovery while maintaining low polarization across both competitor regimes. This supports evaluating response policies by trajectory stability rather than final trust alone. (iii) \textbf{More voice does not mean better outcomes:} higher SoV does not reliably translate into trust gains; credible actions matter more than high-volume messaging.

Mean performance can obscure rare but severe backfires that dominate crisis cost, so we use BrandEval-Dynamic as a tail-risk stress test. We compare the Strategic Agent with two response policies: a Manipulative policy prompted to maximize visibility and suppress competitors even at the expense of accountability, and a Specious policy that produces plausible but weakly grounded statements without verifiable commitments. We report four scenario-level tail-risk metrics: worst-case BRI, CVaR$_{10}$ for BRI over the bottom 10\% scenarios, worst-case NPS, and the catastrophic failure rate under $BRI_T < -10$, all computed after averaging $K=5$ seeds per scenario. Table~\ref{tab:tail_risk_metrics} shows that the dynamic track exposes visibility-seeking and weakly grounded policies as tail-riskier than the SR-guided reference policy in these stress tests.

\begin{table}[htbp]
\small
\centering
\caption{\textbf{Tail risk analysis across 50 scenarios.} Min BRI and Min NPS are the worst-case end-of-horizon outcomes; CVaR$_{10}$ is the mean of the bottom 10\% $BRI_T$ outcomes; Fail rate is the proportion of scenarios with $BRI_T < -10$.}
\setlength{\tabcolsep}{6pt}
\begin{tabular}{lcccc}
\toprule
\textbf{Policy} &
\makecell{\textbf{Min}\\\textbf{BRI$\uparrow$}} &
\makecell{\textbf{CVaR$_{10}$}\\\textbf{BRI$\uparrow$}} &
\makecell{\textbf{Min}\\\textbf{NPS$\uparrow$}} &
\makecell{\textbf{Fail}\\\textbf{rate$\downarrow$}} \\
\midrule
Manipulative     & -18.52 & -16.20 & -35.00 & 12\% \\
Specious         & -10.24 &  -9.15 & -22.00 & 4\% \\
Strategic Agent  & \textbf{-5.21} & \textbf{-4.82} & \textbf{-12.00} & \textbf{0\%} \\
\bottomrule
\end{tabular}
\label{tab:tail_risk_metrics}
\end{table}

\subsection{Static and Dynamic Alignment}
\label{sec:analysis-discrepancies}

We next test the central premise of BrandEval: whether static strategic quality predicts dynamic crisis handling. We descriptively quantify this transfer gap with Kendall's $\tau$ between model rankings from BrandEval-Static and directed BrandEval-Dynamic outcomes, reversing the ranking direction for lower-is-better metrics. Static rankings align with trust and polarization outcomes ($\tau=0.67$ for BRI; $\tau=0.82$ for Pol.) but are negatively associated with process-health metrics (ARV: $\tau=-0.33$; CRS: $\tau=-0.24$). This does not imply the static rubric is wrong: BrandEval-Static rewards one-shot diagnostic coverage, whereas BrandEval-Dynamic stresses multi-turn execution under feedback and competition, where over-commitment, oscillation, and evidence-timing drive stability and recovery. We therefore use BrandEval-Static as a diagnostic complement rather than a standalone proxy for long-horizon process health.

\subsection{Benchmark Sanity Checks}
\label{sec:benchmark_validity}
We further assess whether BrandPolis exhibits coarse-grained alignment with selected rise-and-fall patterns and discourse structures observed in real online brand and consumer interactions.
We study three real-world electric vehicle events, a highway-accident crisis that first broke widely on 2025-03-29~\cite{weibo_chinanewsweek_2025_su7_crash_investigation}, a product-recall incident announced on 2025-09-19~\cite{weibo_yicai_2025_safety_recall_editorial}, and a spontaneous-combustion incident on 2025-10-13~\cite{weibo_xiaoxiangchenbao_2025_chengdu_vehicle_fire_rescue}.
We compare simulated outputs against a post-derived reference series under matched operationalization, and compare discourse statistics under matched preprocessing, to check whether the sandbox exhibits coarse alignment with rise--fall patterns and salient topic structure (Table~\ref{tab:sandbox-validity}).

\begin{table*}[htbp]
\centering
\caption{\textbf{Comparisons and ablations on BrandEval-Dynamic.} Results are grouped into a backbone baseline, prompting baselines with the same backbone, and Strategic Agent variants that ablate SR.}
\small
\setlength{\tabcolsep}{3.5pt}
\renewcommand{\arraystretch}{1.15}
\resizebox{\textwidth}{!}{%
\begin{tabular}{l|cc|cc|cc|cc|cc|cc}
\toprule
\multirow{2}{*}{\textbf{Model}} &
\multicolumn{2}{c|}{\textbf{BRI$\uparrow$}} &
\multicolumn{2}{c|}{\textbf{NPS$\uparrow$}} &
\multicolumn{2}{c|}{\textbf{ARV$\downarrow$}} &
\multicolumn{2}{c|}{\textbf{Pol.$\downarrow$}} &
\multicolumn{2}{c|}{\textbf{CRS$\downarrow$}} &
\multicolumn{2}{c}{\textbf{SoV}} \\
\cline{2-13}
& \raisebox{-0.6ex}{\textbf{Single}} & \raisebox{-0.6ex}{\textbf{Dual}}
& \raisebox{-0.6ex}{\textbf{Single}} & \raisebox{-0.6ex}{\textbf{Dual}}
& \raisebox{-0.6ex}{\textbf{Single}} & \raisebox{-0.6ex}{\textbf{Dual}}
& \raisebox{-0.6ex}{\textbf{Single}} & \raisebox{-0.6ex}{\textbf{Dual}}
& \raisebox{-0.6ex}{\textbf{Single}} & \raisebox{-0.6ex}{\textbf{Dual}}
& \raisebox{-0.6ex}{\textbf{Single}} & \raisebox{-0.6ex}{\textbf{Dual}} \\
\midrule

\multicolumn{13}{c}{\textit{Backbone baseline}} \\
Qwen3-8B
& -3.54 & -4.30
& -3.00 & -9.00
& 30.06 & \textbf{29.74}
& \underline{1.98} & \textbf{1.98}
& 16 & 15
& 63.46 & 68.90 \\

\midrule\midrule
\multicolumn{13}{c}{\textit{Prompting baselines (same backbone, different prompting)}} \\
Qwen3-8B + CoT
& -2.98 & -3.88
& \textbf{-1.00} & -4.00
& 29.50 & \underline{30.60}
& 3.92 & 5.82
& 16 & 15
& 68.93 & 69.58 \\

Qwen3-8B + ReAct
& -2.89 & \underline{-3.74}
& -5.00 & -5.00
& 35.16 & 31.95
& \textbf{0.00} & \underline{3.92}
& \textbf{13} & \textbf{12}
& 70.30 & 69.62 \\

Qwen3-8B + Reflexion
& \underline{-2.76} & -3.86
& \underline{-2.00} & \textbf{-1.00}
& 34.38 & 30.72
& \textbf{0.00} & \underline{3.92}
& \underline{14} & \underline{13}
& 68.63 & 69.61 \\

\midrule\midrule
\multicolumn{13}{c}{\textit{Strategic Agent ablations (policy/workflow components)}} \\
Strategic Agent (w/o SR)
& -3.10 & -3.96
& \underline{-2.00} & -3.00
& \underline{28.90} & 31.20
& \underline{1.98} & \textbf{1.98}
& 15 & 14
& 67.65 & 69.30 \\

\rowcolor{gray!15}
Strategic Agent (SR)
& \textbf{-2.36} & \textbf{-2.76}
& \textbf{-1.00} & \underline{-2.00}
& \textbf{27.78} & 31.83
& \underline{1.98} & \textbf{1.98}
& \underline{14} & \underline{13}
& 69.61 & 69.99 \\

\bottomrule
\end{tabular}%
}
\label{tab:ablation}
\end{table*}

\begin{table}[htbp]
\centering

\caption{\textbf{BrandPolis sanity checks.} Macro-level trend alignment and micro-level discourse reproduction across three crisis types.}
\small
\setlength{\tabcolsep}{3pt}
\begin{tabular}{lccc}
\toprule
\textbf{Metric} &
\makecell{\textbf{Highway}\\\textbf{accident}} &
\makecell{\textbf{Product}\\\textbf{recall}} &
\makecell{\textbf{Spontaneous }\\\textbf{combustion}} \\
\midrule
\multicolumn{4}{c}{\textit{Macro trend alignment}} \\
Lagged $r_{\max}$ ($\pm 2$) $\uparrow$ & 0.60 & 0.80 & 0.75 \\
Spearman $\rho$ $\uparrow$             & 0.62 & 0.56 & 0.59 \\
Norm-RMSE $\downarrow$                 & 0.35 & 0.32 & 0.29 \\
DTW (Z-norm) $\downarrow$              & 0.21 & 0.27 & 0.23 \\
90\% band coverage $\uparrow$          & 0.60 & 0.65 & 0.65 \\
\midrule\midrule
\multicolumn{4}{c}{\textit{Micro discourse reproduction}} \\
Topic JSD $\downarrow$                 & 0.18 & 0.21 & 0.17 \\
Topic cosine $\uparrow$                & 0.74 & 0.62 & 0.76 \\
Top-10 overlap $\uparrow$              & 0.60 & 0.70 & 0.70 \\
\bottomrule
\end{tabular}
\label{tab:sandbox-validity}
\end{table}

\paragraph{Macro-level trend alignment.}
We sanity-check whether BrandPolis exhibits coarse alignment with event-level rise--peak--recovery patterns by comparing its simulated reputation trajectory to a post-derived reference series aggregated from observed crisis discussions under a matched operationalization over the same 60-day horizon; we map each round to a 3-day window to align the 20-round episode with this horizon. We compare simulated and proxy trajectories using lag-aligned Pearson correlation within a fixed $\pm 2$ round window, Spearman rank correlation, normalized RMSE after min--max normalization, dynamic time warping (DTW) on Z-normalized trajectories, and 90\% prediction-band coverage. Across events, correlation-based signals are directionally positive, while distance-based comparisons and coverage suggest broadly similar trajectory shapes under the shared operationalization, providing a macro-level sanity check for BrandPolis.

\paragraph{Micro-level discourse reproduction.}
We further check whether BrandPolis generates plausible crisis discourse structure by comparing simulated and observed posts under matched preprocessing and a shared topic space. We evaluate three aspects: how similar the overall topic mix is (Jensen--Shannon divergence (JSD)), how close the topic meanings are (topic-level cosine similarity), and whether top topics match (Top-10 overlap). Across events, the simulated and real discussions show similar topic mixtures and share many top topics, indicating that BrandPolis recovers broad crisis themes and their approximate prominence during crises.

\subsection{Response Strategy Analysis}
We run ablations on BrandEval-Dynamic to disentangle generic prompting from policy and workflow components in Table~\ref{tab:ablation}. CoT improves advocacy but increases polarization under stronger competition, while ReAct and Reflexion often trade lower polarization for weaker trust or higher volatility. In contrast, the SR-guided Strategic Agent improves over the same backbone, increasing $BRI_T$ from $-3.54/-4.30$ to $-2.36/-2.76$ and $NPS_T$ from $-3.00/-9.00$ to $-1.00/-2.00$, while keeping recovery competitive. Removing SR erodes these gains, indicating that the workflow adds value beyond the underlying reference policy by enforcing verifiable goals and feedback-aware plan selection, with additional robustness checks reported in the Appendix.

\section{Conclusion and Discussion}
This paper introduces BrandEval, a two-track benchmark for risk-sensitive product crisis communication that pairs a static stakeholder-aware diagnostic with BrandPolis, a controlled multi-agent stress-test environment. Across proprietary and open-source models, BrandEval shows that static strategic quality is useful but incomplete: single-turn analyses capture part of response quality, yet miss process health failures and tail risks that emerge under stakeholder feedback and competitor actions. To support controlled evaluation, we construct BrandSRD and use Strategic Rationale to instantiate a reference Strategic Agent.

Static and dynamic evaluation serve complementary roles. BrandEval-Static diagnoses stakeholder awareness, evidence use, and recommendation quality, while BrandEval-Dynamic tests how communication decisions reshape trust, risk, and recovery under interaction. This separation helps distinguish plausible analyses from agents that sustain credible crisis handling over time, suggesting that effective crisis communication agents should integrate verifiable goal setting, persistent stakeholder tracking, and future-aware action selection.

\section*{Limitations}
\paragraph{Simulator abstraction.}
BrandPolis is a controlled stress-test environment, not a real-world reputation forecaster. It uses a simplified interaction structure, simplified CBBE/TPB updates, a single-channel platform abstraction, manually specified hyperparameters, and no explicit social-graph tie strengths. Our event-level trend checks and sensitivity analyses should therefore be interpreted as sanity checks for the sandbox rather than definitive external-validity evidence.

\paragraph{Data and platform scope.}
BrandSRD and BrandEval are instantiated using Chinese social platforms and selected consumer domains. The benchmark structure is not tied to one language or market, but applying it to other markets, languages, channels, or regulatory contexts will require re-calibration.

\paragraph{Silver labels.}
BrandSRD is semi-automatically constructed: preference labels are distilled with model assistance and then filtered and audited by crowd workers. They remain silver-standard signals and may reflect generator, annotator, or platform-specific biases.

\paragraph{LLM judge proxy.}
BrandEval-Static uses a small three-model judging panel audited against limited human ratings. Its scores should be interpreted as scalable diagnostic proxies rather than standalone measures of real-world robustness, and should be used together with BrandEval-Dynamic to assess long-horizon behavior under interaction.

\section*{Ethical Considerations}
All data in BrandSRD are derived from publicly accessible posts on Weibo, Douyin, and Xiaohongshu, supplemented with public news reports and official announcements for event context. We remove or rewrite identifying information and release only de-identified, self-contained scenario briefs and aggregated annotations rather than user-identifiable records. We will release only the de-identified BrandSRD records and will not release any original social-media posts. The counterfactual response plans are generated via commercial LLM APIs solely to construct a risk-aware evaluation and training resource. BrandSRD, BrandEval, and BrandPolis are intended for controlled evaluation, failure analysis, and accountability-oriented policy comparison, not for operational persuasion or deployment-ready crisis management. We explicitly prohibit using these artifacts to generate misleading crisis messaging, suppress criticism, evade accountability, or manipulate public opinion; the benchmark design further penalizes unverifiable claims and manipulative tactics.

\section*{Acknowledgments}

This work was supported by the National Natural Science Foundation of China under grant No. 62606277.

\bibliography{refer}

\appendix
\clearpage

\section*{Appendix}
\section{Reference Strategic Agent}
\label{app:reference_agent}

This section describes how we instantiate the reference Strategic Agent used in BrandEval-Dynamic. The goal is to obtain a fixed and reproducible reference policy for studying structured crisis-response reasoning. All baseline LLMs can be evaluated in BrandEval-Dynamic without this training procedure.

\subsection{Training Data Format}
\label{app:alignment_data_format}

Each BrandSRD instance is represented as
\begin{equation}
z_i = (c_i, a_i^{\mathrm{obs}}, \mathcal{A}_i^{\mathrm{cf}}, y_i^{+}, {y}_i^{-}, r_i),
\end{equation}
where $c_i$ is the crisis context, $a_i^{\mathrm{obs}}$ is the observed brand response, $\mathcal{A}_i^{\mathrm{cf}}$ denotes generated counterfactual response options, $y_i^{+}$ is the preferred response, ${y}_i^{-}$ is the set of non-preferred alternatives, and $r_i$ is the rationale under the stakeholder rubric.
The input prompt $x_i$ includes the crisis context, stakeholder information, and formatting instructions.
The target follows the Strategic Rationale format, including a verifiable goal, stakeholder-state summary, candidate plans, and a final selected response.
For deployment analysis, we log one SR record per Strategic Agent decision in BrandPolis; only the selected plan is posted to the environment.

\Needspace{12\baselineskip}
\begin{FormatBox}{SR record schema JSON template}
\texttt{\{}\\
\texttt{\ \ "round\_id": "...",}\\
\texttt{\ \ "observation\_summary": "...",}\\
\texttt{\ \ "goal": "...",}\\
\texttt{\ \ "stakeholder\_summary": "...",}\\
\texttt{\ \ "candidate\_plans": [ "...", "...", "..." ],}\\
\texttt{\ \ "selected\_plan": "...",}\\
\texttt{\ \ "final\_action\_content": "...",}\\
\texttt{\ \ "claimed\_evidence": "..."}\\
\texttt{\}}\\[6pt]
\end{FormatBox}

\subsection{Training Objectives}
\label{app:training_objectives}
We initialize from Qwen3-8B and align it on BrandSRD to obtain the policy model $\pi_\theta$ for the reference Strategic Agent. The policy is deployed with the SR workflow, which structures each round into goal setting, stakeholder-state tracking, candidate planning, and final action selection. Training follows a two-stage pipeline.

\subsubsection{Stage A: Format Bootstrapping} \label{app:stage_a_format}
Stage A bootstraps parseable and evidence-aware SR-style outputs using an iterative DPO objective. The goal is to constrain the model to produce parseable, structured strategic analyses and to construct learnable preference signals for open-ended tasks without explicit ground truth. Given an input scenario $x$ and its evaluation-aligned prompt $\rho(x)$, we sample a candidate set $\{y_k\}_{k=1}^{K}$ and apply a verifiable gate $G(x,y)\in\{0,1\}$ to filter out unparsable or schema-violating outputs. For candidates that pass the gate, a frozen self-judge produces a normalized reward $R(x,y)\in[0,1]$ with a hard evidence constraint: evidence spans must be substrings of the original prompt text; otherwise we set $R(x,y)=0$. We then select, among gated candidates,
\begin{equation}
\label{eq:stageA_select_en}
\begin{aligned}
y^{+}&=\arg\max_{y} R(x,y),\\
y^{-}&=\arg\min_{y} R(x,y),
\end{aligned}
\end{equation}
to form a preference triple $(x,y^{+},y^{-})$. Stage A optimizes an Iterative DPO objective anchored by a reference model $\pi_{\text{ref}}$ to constrain update magnitude:
\begin{equation}
\label{eq:stageA_loss_en}
\begin{adjustbox}{max width=0.85\columnwidth}
$
\begin{aligned}
\mathcal{L}_{A}(\theta)
&= -\mathbb{E}_{(x,y^{+},y^{-})}\Big[
\log \sigma\big(\beta(\Delta_\theta-\Delta_{\text{ref}})\big)
\Big],
\end{aligned}
$
\end{adjustbox}
\end{equation}
where
\begin{equation}
\label{eq:stageA_delta_en}
\begin{aligned}
\Delta_\theta
&=\log \pi_\theta(y^{+}\mid x)-\log \pi_\theta(y^{-}\mid x), \\
\Delta_{\text{ref}}
&=\log \pi_{\text{ref}}(y^{+}\mid x)-\log \pi_{\text{ref}}(y^{-}\mid x),
\end{aligned}
\end{equation}
and $\beta$ is a strength coefficient. Stage A only bootstraps parseable and evidence-aware SR-style outputs, and the resulting adapter initializes Stage B.

\subsubsection{Stage B: BrandSRD Alignment} 
\label{app:stage_b_preference}
Stage B aligns the bootstrapped model on BrandSRD preference annotations using an ORPO-style objective. Each BrandSRD instance provides a strategic context, candidate responses, a preferred response, dispreferred alternatives, and a rationale. We form preference triples $(x,y^{+},y^{-})$, where $y^{+}$ is the annotated preferred response and $y^{-}$ is sampled from the dispreferred candidate set. We optimize an objective that combines a preference log-odds term with conditional cross-entropy on $y^{+}$:
\begin{equation}
\label{eq:stageB_loss_en}
\begin{aligned}
\mathcal{L}_{B}(\theta)
&=
-\log \sigma\!\Bigg(
\beta\Big(
\log \pi_\theta(y^{+}\mid x)\\
&\quad-\log \pi_\theta(y^{-}\mid x)
\Big)
\Bigg)
\\
&\quad
+ \lambda\,\bigl(-\log \pi_\theta(y^{+}\mid x)\bigr),
\end{aligned}
\end{equation}
where $\beta$ controls the preference term and $\lambda$ weights the supervised term. The Stage B-aligned policy serves as the language policy model of Strategic Agent in BrandPolis.

\subsection{Strategic Rationale Deployment}
\label{app:sr_deployment}

At deployment time, the aligned policy is wrapped with the Strategic Rationale workflow.
At each BrandPolis round $t$, the focal agent receives a market brief $S_t$ containing platform-visible aggregates, dominant topics, recent public reactions, and its own published commitments.
The agent then produces an action through four structured steps:
\begin{enumerate}[leftmargin=*, itemsep=1pt, topsep=2pt]
    \item \textbf{Goal verification:} state a verifiable communication goal that balances business needs with social responsibility.
    \item \textbf{Stakeholder tracking:} summarize the current state of consumers, public and media actors, and competitors based only on observable information.
    \item \textbf{Candidate planning:} draft several publishable response candidates with explicit assumptions, evidence use, and expected risks.
    \item \textbf{Future-aware selection:} select the response that best balances credibility, de-escalation, recovery potential, and downside risk.
\end{enumerate}
The selected response is published into BrandPolis and becomes part of the public message log for the next round.
BrandPolis then updates consumer states, competitor observations, and platform-visible market briefs according to the environment dynamics in Section~\ref{sec:brandpolis}.

\subsection{Role of the Reference Agent}
\label{app:reference_agent_role}

The SR-based Strategic Agent is a controlled policy implementation, not a benchmark requirement.
BrandEval-Dynamic can evaluate any model or agent that produces a publishable crisis-response action from the current market brief.
The reference agent provides a reproducible policy for stress-testing BrandPolis and for comparing against zero-shot LLM baselines.
It also supports analysis of whether goal verification, stakeholder tracking, and future-aware selection improve long-horizon trust and stability.
Thus, conclusions about BrandEval rely on the benchmark protocol and dynamic outcomes, not on assuming that SR is the only valid policy design.

\section{Benchmark Construction and Audits}
\label{app:benchmark_audits}

\subsection{Scenario Construction}
\label{app:scenario_construction}
BrandEval-Static scenarios are constructed by fictionalizing a held-out prototype pool with Gemini-2.5-Pro,
followed by human screening for commercial plausibility and removal of any identifying details.
Fictionalization preserves the underlying strategic trade-offs and multi-stakeholder tensions,
while ensuring that no real-world entities, individuals, handles, or URLs remain.
Disjointness from the BrandSRD training split is enforced by (i) pre-reserving the evaluation pool before training-split construction and (ii) filtering with prototype-level IDs and metadata to prevent overlap.

\subsection{BrandEval-Static Rubric}
\label{app:static_rubric}
BrandEval-Static is scored using a structured rubric inspired by stakeholder theory.
The rubric consists of four dimensions, each scored on a 1--10 scale (higher is better). These dimensions are designed to capture stakeholder diagnosis, evidence use, competitive reasoning, and strategic synthesis, which are later reflected in BrandPolis through perception and intent-related state channels.
Table~\ref{tab:static_rubric} summarizes the operational definitions.

\begin{table*}[t]
\centering
\small
\caption{\textbf{BrandEval-Static rubric.} Four dimensions scored on a 1--10 scale.}
\setlength{\tabcolsep}{6pt}
\renewcommand{\arraystretch}{1.1}
\begin{tabular}{p{0.20\textwidth}p{0.74\textwidth}}
\toprule
\textbf{Dimension} & \textbf{Operational definition} \\
\midrule
Consumer \& Brand &
Accurate diagnosis of consumer concerns; concrete brand actions; credible evidence (data, timelines, accountability);
explicit linkage to brand equity dimensions (competence/value/emotion) when appropriate. \\

Public \& Media &
Anticipation of public/media narratives; risk-aware messaging; clarity on responsibility/safety; preventive escalation control;
alignment with transparency and verifiability. \\

Competitor &
Recognition of competitive dynamics (e.g., opportunistic attacks, SoV shifts); defensive/offensive counter-moves;
avoids self-inflicted openings. \\

Strategic Synthesis &
Coherent trade-offs across stakeholders; prioritized goals; operational plan with milestones; avoids generic PR;
actionability and consistency with constraints. \\
\bottomrule
\end{tabular}
\label{tab:static_rubric}
\end{table*}

\subsection{Dynamic Scenario Selection}
\label{app:dynamic_scenario_selection}
For BrandEval-Dynamic, we select 50 scenarios from the held-out pool via stratified sampling over
(i) dominant topic clusters and (ii) risk profiles (e.g., safety, governance, product quality, pricing).
Risk profiles are annotated during scenario curation and used only to balance evaluation coverage.

\subsection{Judge System and Bias Diagnostics}
\label{app:judge_bias}

We audit the BrandEval-Static judge panel from two complementary angles: robustness to superficial output changes and sensitivity to evidence grounding. The first audit tests whether scores are driven by presentation artifacts, while the second tests whether evidence-sensitive judging can separate verifiable strategies from specious but fluent responses.

\subsubsection{Robustness to Surface Changes} \label{app:judge_surface_changes}
As a targeted robustness check, we conduct sensitivity analyses with style and structural perturbations. We select a random subset of 5 scenarios and perturb the model outputs before feeding them to the judge panel.

\paragraph{Perturbation methods.}
We apply two types of perturbations while preserving the substantive content of each output:
\begin{itemize}
    \item \textbf{Style perturbation:} We rewrite the agent's analysis using two distinct styles, ``Flowery'' with more elaborate language and ``Concise'' with simpler and shorter language, while preserving the original insights and semantic content.
    \item \textbf{Structure perturbation:} We shuffle the order of the four analysis sections, namely Customer, Public, Competitor, and Strategic Synthesis, in the JSON input to test whether the judge score is sensitive to presentation order.
\end{itemize}

\paragraph{Results.}
Across all four evaluation dimensions, the mean absolute score change is 0.17 for the flowery style perturbation, 0.42 for the concise style perturbation, and 0.25 for section-order shuffling. The maximum absolute change is 1 point in all three settings, and the 90th percentile remains at or below 1.0. These results suggest that the scoring protocol is not primarily driven by formatting, section order, or superficial linguistic style.

\subsubsection{Evidence Sensitive Judge Audit}
\label{app:audit-judge}

We further evaluate whether the scoring protocol can distinguish verifiable strategies from specious but fluent responses. We construct paired comparisons between evidence-grounded strategies and specious alternatives that contain plausible rhetoric but lack adequate support or accountability. To test whether the evidence-sensitive judge is merely uniformly harsher, we evaluate its paired discriminability between \textbf{Verifiable} and \textbf{Specious} strategies on matched stress scenarios.

For each scenario $i$, we compute the paired gap
\begin{equation}
\Delta_i = \mathrm{Score}_i^{\mathrm{Verifiable}} - \mathrm{Score}_i^{\mathrm{Specious}},
\end{equation}
and summarize the mean gap $\overline{\Delta}$, a bootstrap 95\% confidence interval over $\{\Delta_i\}_{i=1}^{N}$ with 1{,}000 resamples, the win rate $\Pr(\Delta_i>0)$, and paired Cohen's $d = \overline{\Delta}/s_{\Delta}$, where $s_{\Delta}$ is the standard deviation of paired differences.

\begin{table}[!htbp]
\centering
\scriptsize
\caption{\textbf{Audit Judge paired discriminability} ($N=9$). The evidence-sensitive judge better separates Verifiable and Specious strategies under matched stress scenarios.}
\label{tab:audit_discriminability}
\setlength{\tabcolsep}{2pt}
\renewcommand{\arraystretch}{1.06}
\resizebox{0.95\linewidth}{!}{%
\begin{tabular}{lcccc}
\toprule
\textbf{Judge} & \textbf{Mean $\overline{\Delta}$} & \textbf{95\% CI} & \textbf{Win(\%)} & \textbf{Paired $d$} \\
\midrule
Standard & $+0.44$ & $[-0.33, 1.33]$ & $44$ & $0.35$ \\
Evidence-sensitive & $\mathbf{+3.11}$ & $\mathbf{[1.67, 4.56]}$ & $\mathbf{89}$ & $\mathbf{1.34}$ \\
\bottomrule
\end{tabular}%
}
\end{table}
The evidence-sensitive judge is selectively specific: it increases separation primarily when evidence quality differs. This indicates that the audit protocol is sensitive to evidence grounding rather than only surface fluency.

\subsubsection{Rubric Reliability}
\label{app:expC-rubric}

\paragraph{Internal consistency.}
We evaluate internal consistency across the rubric's dimensions using Cronbach's $\alpha$:
\begin{equation}
\alpha
=
\frac{k}{k-1}
\left(
1 - \frac{\sum_{j=1}^{k}\sigma_j^2}{\sigma_{\text{total}}^2}
\right),
\label{eq:cronbach_alpha}
\end{equation}
where $k$ is the number of rubric dimensions, $\sigma_j^2$ is the variance of dimension $j$, and $\sigma_{\text{total}}^2$ is the variance of the summed score.
We obtain $\alpha = 0.9012$, indicating high internal consistency among rubric dimensions.

\section{BrandPolis Mechanics and Metrics}
\label{app:brandpolis_details}

\subsection{BrandPolis Update Equations}
\label{app:brandpolis_equations}

This section provides implementation details for the platform exposure, retention, and evidence-grounding mechanisms used in BrandEval-Dynamic.

\subsubsection{Exposure and Retention}
\label{app:brandpolis-env}

Official posts are treated as broadcast signals with $P_{\text{push}}(m,t)=1$. 
For UGC and other non-broadcast messages, the platform samples feed candidates according to $P_{\text{push}}(m,t)$ and retains at most $C$ messages per consumer per round. Broadcast items are prioritized, and remaining slots are filled with sampled non-broadcast items.

For a non-broadcast message $m$ from sender $s$ on topic $k$ at round $t$, exposure is:
\begin{equation}
\label{eq:app-push}
\begin{adjustbox}{max width=0.89\columnwidth}
$
\begin{aligned}
P_{\text{push}}(m,t)
&= \operatorname{clip}_{[0,1]}\Bigl(
P_{\text{base}}\Bigl[
1 + c_1 \ln\!\bigl(I_s(t)+1\bigr)
\\
&\qquad\qquad
+ c_2 \ln\!\bigl(H_k(t)+1\bigr)
\Bigr]
\Bigr),
\end{aligned}
$
\end{adjustbox}
\end{equation}
where $P_{\text{base}}$ is the global base rate, $I_s(t)$ is the sender influence score, and $H_k(t)$ is topic heat computed from recent message volume and interactions.

The probability that consumer $u$ processes message $m$ is:
\begin{equation}
\label{eq:app-ret}
P_{\mathrm{ret}}(m,u,t)
=
P_{\mathrm{push}}(m,t)\, s_u\, a_{m,u},
\end{equation}
where $s_u\sim\mathcal{U}(0,1)$ is the information-retention susceptibility, and $a_{m,u}\in[0,1]$ denotes the stance-alignment factor between message $m$ and consumer $u$.

The observation builder uses only the public message log and engagement counters. It does not read latent state tensors used for computing BRI, NPS, Pol., ARV, or CRS.

\subsubsection{Action Grounding Mechanism}
\label{app:action-grounding-mech}

BrandPolis maintains an evidence store $\mathcal{E}^{(t)}$. At each round, the strategic agent may output \texttt{claimed\_evidence} to indicate that an official action is grounded in a checkable artifact, such as logs, inspection reports, or third-party findings. The evidence-consistency gate $\gamma^{(t)}\in[0,1]$ checks whether \texttt{claimed\_evidence} matches an identifier in $\mathcal{E}^{(t)}$. Unsupported evidence references are therefore attenuated through the belief update in Eq.~\eqref{eq:app-ema}.

\subsubsection{State Update and Mapping}
\label{app:math}
Each consumer maintains a latent state vector $\mathbf{V}_{u}^{(t)}$ for the focal brand,
initialized at $t=1$ and updated for $t=2,\ldots,T$ via an exponential moving average
with bounded nonlinearity:
\begin{equation}
\label{eq:app-ema}
\mathbf{V}_{u}^{(t)}
=(1-\lambda_u)\,\mathbf{V}_{u}^{(t-1)}
+ \lambda_u\gamma^{(t)}\tanh\!\big(\mathbf{X}_{u}^{(t)}\big).
\end{equation}
where $\mathbf{X}_{u}^{(t)}$ is the message-derived impact signal and $\lambda_u$ is a consumer-specific update rate.

To ground the update dynamics, we fit a text-to-state mapping on a separate human-rated subset.
Concretely, we extract lightweight textual features from each message and fit multivariate regression models to predict targets aligned with CBBE and TPB, including the three perception dimensions and the two intention dimensions.
The regression predictions are used as the message-derived impact signal $\mathbf{X}_{u,j}^{(t)}$ in Eq.~\eqref{eq:app-ema}, so that consumer states evolve as a first-order EMA driven by observed text.
We treat this mapping as an explicit and reproducible modeling choice that keeps the simulator transparent and auditable, while recognizing that the latent states are internal proxies and require independent validation to support real-world interpretation.

\subsection{BrandEval-Dynamic Metrics}
\label{app:dynamic_metrics}
BrandEval-Dynamic reports six metrics. BRI and NPS are reported on $[-100,100]$ point scales; SoV and Polarization are reported in percent; ARV summarizes normalized reputation volatility; and CRS is reported in timesteps.

\paragraph{Brand Reputation Index (BRI).}
\label{app:bri}
BRI captures an aggregate consumer perception signal over three CBBE-aligned dimensions: competence belief (CB), value congruence (VC), and emotional connection (EC).
Let $\overline{CB}_t, \overline{VC}_t, \overline{EC}_t \in [-1,1]$ be population averages at timestep $t$.
We first define a raw latent reputation signal:
\begin{equation}
\label{eq:bri_raw}
b_t = w_{\text{cap}} \cdot \overline{CB}_t + w_{\text{val}} \cdot \overline{VC}_t + w_{\text{emo}} \cdot \overline{EC}_t,
\end{equation}
and define the reported Brand Reputation Index (BRI) as:
\begin{equation}
\label{eq:bri}
BRI_t = 100 \cdot b_t \in [-100,100].
\end{equation}
We use uniform weights to avoid introducing arbitrary emphasis: $w_{\text{cap}} = w_{\text{val}} = w_{\text{emo}} = 1/3$.

\paragraph{Net Promoter Score (NPS).}
\label{app:nps}
Let $\pi^{\text{pro}}_t \in [0,1]$ be the share of Promoters (Advocacy Intent $>0.7$) and
$\pi^{\text{det}}_t \in [0,1]$ be the share of Detractors (Advocacy Intent $<0.3$).
We define the reported NPS on the standard point scale:
\begin{equation}
\label{eq:nps}
NPS_t = 100 \cdot (\pi^{\text{pro}}_t - \pi^{\text{det}}_t) \in [-100,100].
\end{equation}

\paragraph{Share of Voice (SoV).}
\label{app:sov}
In BrandPolis, we operationalize SoV as a topic-focused share of discussion within the simulated market:
\begin{equation}
\label{eq:sov}
SoV_{\text{brand},t} = 100 \cdot \frac{V_{\text{brand},t}}{V_{\text{total},t}} \in [0,100],
\end{equation}
where $V_{\text{brand},t}$ counts all brand-related messages in the sandbox at round $t$ (official brand posts plus consumer UGC responding to the focal brand or event), and $V_{\text{total},t}$ counts all messages generated within the sandbox at round $t$.

\paragraph{Polarization Index.}
\label{app:pol}
We use a bounded group-based polarization score over supporters ($S$), neutrals ($N$), and opponents ($O$).
Let $G=3$, with group shares $s_{g,t}$ at round $t$ (summing to 1), and group locations $\mu_g$.
We set $\mu_S=1$, $\mu_N=0$, and $\mu_O=-1$.
We compute a raw polarization score using a pairwise-distance form:
\begin{equation}
\label{eq:pol_raw}
p_t
=
\sum_{g=1}^{G}\sum_{h=1}^{G} s_{g,t}s_{h,t}\,|\mu_g-\mu_h|,
\end{equation}
where $p_t\in[0,1]$ under this discrete encoding.
We report the Polarization Index on a percent scale:
\begin{equation}
\label{eq:pol}
Pol_t = 100 \cdot p_t \in [0,100].
\end{equation}

Because polarization uses three discrete stance groups with a finite consumer population, the reported Pol values can appear quantized. We therefore treat Pol as a diagnostic for regime shifts rather than as a fine-grained continuous signal.

\paragraph{Average Reputation Volatility (ARV).}
\label{app:ARV}
ARV summarizes \emph{process instability} largely independent of the absolute reputation level.

For an episode, let $\{BRI_t\}_{t=1}^{T}$ be the reported reputation trajectory (Eq.~\ref{eq:bri}).
We first apply per-episode min--max normalization:
\begin{equation}
\label{eq:rv_norm}
\begin{adjustbox}{width=0.85\columnwidth}
$
\displaystyle
\widetilde{BRI}_t =
\frac{BRI_t - \min\limits_{1\le \tau \le T} BRI_\tau}
{\max\limits_{1\le \tau \le T} BRI_\tau - \min\limits_{1\le \tau \le T} BRI_\tau + 10^{-12}}.
$
\end{adjustbox}
\end{equation}
$\widetilde{BRI}_t \in [0,1]$.
This normalization makes ARV comparable across scenarios with different absolute severity by focusing on \emph{trajectory shape volatility} rather than raw amplitude.

We then define Average Reputation Volatility as the standard deviation of the normalized trajectory:
\begin{equation}
\label{eq:ARV}
\begin{aligned}
ARV &= 100 \cdot \sqrt{\frac{1}{T} \sum_{t=1}^{T} (\widetilde{BRI}_t - \overline{\widetilde{BRI}})^2}, \\
\overline{\widetilde{BRI}} &= \frac{1}{T}\sum_{t=1}^{T} \widetilde{BRI}_t .
\end{aligned}
\end{equation}

ARV is reported on a $[0,100]$-like scale.

\paragraph{Crisis Recovery Speed (CRS).}
\label{app:crs}
CRS measures how quickly the reputation trajectory \emph{returns to the episode entry level} after the episode bottom.
Let $t_{\text{bottom}}=\arg\min_{1\le t \le T} BRI_t$.
We define the recovery time as the first timestep after $t_{\text{bottom}}$ such that the reputation reaches the entry level:
\begin{equation}
\begin{adjustbox}{width=0.85\columnwidth}
$
\displaystyle
t_{\text{recover}}=\min\{t>t_{\text{bottom}}: BRI_t \ge BRI_{1}\}.
$
\end{adjustbox}
\end{equation}
We then define CRS (in timesteps) as:
\begin{equation}
\label{eq:crs}
\begin{adjustbox}{width=0.85\columnwidth}
$
\displaystyle
CRS =
\begin{cases}
t_{\text{recover}} - t_{\text{bottom}}, & \text{recovery},\\
(T+1)-t_{\text{bottom}}, & \text{no recovery.}
\end{cases}
$
\end{adjustbox}
\end{equation}
Lower CRS indicates faster recovery under the same horizon and unit scale.

\subsection{Sanity-check Computation Details}
\label{app:validity_metrics}

This section defines the sanity-check metrics reported in Table~\ref{tab:sandbox-validity} and the preprocessing used to align real and simulated series.

\paragraph{Proxy Reputation Trajectory.}
\label{app:real-bri}
Within each 3-day window, we rate observed social posts on three CBBE-aligned dimensions and aggregate them into a proxy reputation score with uniform weights (Eq.~\eqref{eq:bri_raw}).
We report the proxy on the same point scale as the simulator via $BRI = 100 \cdot b$ (Eq.~\eqref{eq:bri}) for comparability of trends across the 20 windows.
The window-level proxy score is computed as the mean over all posts in that window.

\paragraph{Macro Trend Metrics.}
\label{app:macro-metrics}
Let $S_t$ denote the simulated BRI trajectory and $R_t$ denote a proxy reputation trajectory constructed from observed posts under a matched rubric for sanity checking at the same 20-window resolution.

\paragraph{Lag-aligned Pearson correlation.}
To account for plausible diffusion delays between simulation and real-world reactions, we compute Pearson correlation under integer lags $k \in \{-2,-1,0,1,2\}$ and report
$r_{\max}=\max_k \mathrm{corr}(S_t, R_{t+k})$.

\paragraph{Spearman rank correlation.}
We compute Spearman $\rho$ between $S_t$ and $R_t$ (after applying the best lag found above) to measure agreement in rise-and-fall ordering.

\paragraph{Normalized RMSE.}
We min--max normalize each trajectory into $[0,1]$ independently and report the RMSE in the normalized space.

\paragraph{DTW on Z-normalized trajectories.}
We compute DTW distance after Z-normalizing each trajectory to zero mean and unit variance, so DTW reflects shape similarity rather than absolute scale.

\paragraph{90\% prediction-band coverage.}
For each real-world window $t$, we compute the empirical standard deviation of post-level BRI scores within that window and form a 90\% band around the real mean. We then measure the fraction of windows whose simulated value falls inside the corresponding band. To compare on a common scale, we first linearly rescale the simulated series over the 20 windows to match the real series mean and standard deviation, then evaluate coverage.

\paragraph{Discourse Comparison.}
\label{app:topic-pipeline}
To compare discourse structure, we fit a topic model on the union of real and simulated texts after the same preprocessing. We represent texts using character-level n-gram features and train an LDA model with $k=20$ topics. We then compute three metrics in the shared topic space: Jensen--Shannon divergence between the two topic distributions, topic-level semantic cosine similarity, and Top-10 topic overlap.

\paragraph{Diffusion Signal Checks.}
\label{app:diffusion-sanity}
To reduce reliance on rubric-aligned reputation proxies, we additionally sanity-check whether BrandPolis exhibits coarse-grained alignment with \emph{macro diffusion dynamics} that do not share the CBBE operationalization. Concretely, we compare trajectories from real-world data and from simulation for (i) discussion volume (raw count), (ii) topic heat (interaction-weighted volume, using platform-visible engagement such as repost/comment/like counts; when engagement is unavailable for a window, we fall back to raw volume), and (iii) Share of Voice (SoV) over the same 60-day horizon aligned to Day 0 for each event. These signals are computed from counts and exposure statistics rather than perception rubrics, and therefore serve as an orthogonal check on the simulator’s event-level dynamics.

\paragraph{Alignment metrics.}
For each signal $y(t)$, we compute: (1) Pearson correlation ($r$) between real and simulated series after aligning the horizon to the same 20 windows; (2) peak-time error $\Delta t_{\mathrm{peak}}$, the difference (in windows) between the argmax of the two series; and (3) post-peak exponential decay-rate difference $\Delta \kappa$, obtained by fitting $\log y(t)$ on the post-peak segment and comparing slopes. We report these metrics per event and average them across events for a compact summary.

\begin{table}[t]
\centering
\small
\caption{\textbf{Diffusion-signal alignment.} Metrics are averaged over three real-world crisis events.}
\setlength{\tabcolsep}{4pt}
\renewcommand{\arraystretch}{1.08}
\begin{tabular}{lccc}
\toprule
\textbf{Signal} & \textbf{Corr. ($r$) $\uparrow$} & $\boldsymbol{|\Delta t_{\mathrm{peak}}| \downarrow}$ & $\boldsymbol{|\Delta \kappa| \downarrow}$ \\
\midrule
Volume & 0.56 & 0.3 & 0.48 \\
Heat   & 0.59 & 0.4 & 0.52 \\
SoV    & -0.07 & 9.7 & 0.72 \\
\bottomrule
\end{tabular}
\label{tab:diffusion_sanity}
\end{table}

Volume and Heat show moderate alignment and small peak-time errors, providing an orthogonal macro-level sanity check beyond rubric-aligned reputation proxies. SoV remains inconclusive because real-world official voice is sparse and platform-mediated.

\section{Additional Results and Validity Checks}
\label{app:additional_results}

\subsection{Supplementary Cross-Domain Check}
\label{app:cross_domain_generalization}

The main BrandEval-Dynamic evaluation uses a primary held-out pool for controlled comparison. Since BrandSRD covers multiple product domains, we further conduct a supplementary dynamic check on 30 additional held-out scenarios from four product domains not used in the main dynamic evaluation: smartphones, smart door locks, refrigerators, and speakers. These scenarios are used only for this supplementary check. We compare the SR-guided Strategic Agent with the strongest prompting baseline, Qwen3-8B + CoT, using the same BrandPolis protocol with $T=20$ rounds and $K=5$ random seeds per scenario.

\begin{table}[!htbp]
\centering
\scriptsize
\caption{\textbf{Supplementary cross-domain dynamic check.} Results are averaged over 30 additional held-out scenarios from four product domains with $K=5$ seeds per scenario.}
\label{tab:cross_domain_generalization}
\setlength{\tabcolsep}{2.5pt}
\renewcommand{\arraystretch}{1.06}
\resizebox{0.95\linewidth}{!}{%
\begin{tabular}{lcccc}
\toprule
\textbf{Model / Policy} & \textbf{BRI $\uparrow$} & \textbf{ARV $\downarrow$} & \textbf{CRS $\downarrow$} & \textbf{Fail rate $\downarrow$} \\
\midrule
Qwen3-8B + CoT & $-4.85$ & $32.4$ & $16$ & $6.7\%$ \\
Strategic Agent & $\mathbf{-2.60}$ & $\mathbf{28.1}$ & $\mathbf{13}$ & $\mathbf{0.0\%}$ \\
\bottomrule
\end{tabular}%
}
\end{table}

This supplementary check suggests that the process-health pattern observed in the main dynamic evaluation is not solely tied to the primary held-out pool. Compared with Qwen3-8B + CoT, the Strategic Agent obtains a higher end-of-horizon reputation score, lower trajectory volatility, faster recovery, and fewer tail failures.

\subsection{Sandbox Hyperparameter Sensitivity}
\label{app:hparam_sensitivity}

We also conduct one-at-a-time (OAT) sensitivity analysis for key BrandPolis hyperparameters. For each row, one parameter is varied across five levels while all other parameters remain fixed.

\begin{table*}[t]
 \centering
\caption{\textbf{OAT sensitivity of sandbox hyperparameters.} One parameter is varied per row across five levels, with all others fixed. Values are mean $\pm$ std over seeds.}
 \small
 \setlength{\tabcolsep}{3pt}
 \renewcommand{\arraystretch}{1.10}
 \begin{adjustbox}{width=\textwidth}
 \begin{tabular}{@{}l l c c c c c c@{}}
 \toprule
 \textbf{Parameter} & \textbf{Tested levels (OAT)} & \textbf{BRI$\uparrow$} & \textbf{ARV$\downarrow$} & \textbf{NPS$\uparrow$} & \textbf{Pol.$\downarrow$} & \textbf{CRS$\downarrow$} & \textbf{SoV} \\
 \midrule
 \texttt{base\_p} & 0.04, 0.07, \textbf{0.10}, 0.13, 0.16 & $-2.80 \pm 0.40$ & $35.5 \pm 4.5$ & $-2.10 \pm 0.50$ & $2.10 \pm 0.50$ & $14.5 \pm 0.5$ & $70.5 \pm 6.0$ \\
 \texttt{virality} & 0.00, 0.05, \textbf{0.10}, 0.15, 0.20 & $-3.10 \pm 0.60$ & $40.0 \pm 12.0$ & $-2.80 \pm 1.20$ & $4.50 \pm 2.20$ & $14.5 \pm 1.5$ & $70.0 \pm 10.0$ \\
 \texttt{topic\_coeff} & 0.00, 0.05, \textbf{0.10}, 0.15, 0.20 & $-3.05 \pm 0.20$ & $32.0 \pm 2.0$ & $-1.90 \pm 0.30$ & $2.00 \pm 0.30$ & $14.0 \pm 0.2$ & $70.0 \pm 2.0$ \\
 \texttt{feed\_cap} & 10, 15, \textbf{20}, 30, 40 & $-2.90 \pm 0.30$ & $30.0 \pm 3.0$ & $-2.00 \pm 0.40$ & $1.95 \pm 0.40$ & $14.5 \pm 0.5$ & $70.0 \pm 3.5$ \\
 \texttt{noise\_delta} & 0.00, 0.05, \textbf{0.10}, 0.15, 0.20 & $-3.15 \pm 0.50$ & $33.5 \pm 1.5$ & $-2.20 \pm 0.60$ & $5.80 \pm 3.50$ & $15.0 \pm 1.0$ & $70.0 \pm 0.5$ \\
 \texttt{decay\_lambda} & 0.50, 0.75, \textbf{1.00}, 1.25, 1.50 & $-3.50 \pm 1.20$ & $31.0 \pm 2.5$ & $-2.50 \pm 0.80$ & $2.05 \pm 0.60$ & $14.0 \pm 0.1$ & $70.0 \pm 1.0$ \\
 \texttt{comp\_pressure} & 0.25, 0.50, \textbf{1.00}, 1.50, 2.00 & $-3.20 \pm 0.40$ & $28.0 \pm 10.0$ & $-2.30 \pm 0.70$ & $2.10 \pm 0.40$ & $14.5 \pm 0.5$ & $69.0 \pm 12.0$ \\
 \bottomrule
 \end{tabular}
 \label{tab:hparam_sensitivity}
 \end{adjustbox}
\end{table*}
\subsection{Robustness and Sensitivity}
\label{app:robustness} 

This section complements the main evaluation setting ($K=5$ seeds per model and scenario pair) with additional robustness checks.
We report one-at-a-time sandbox hyperparameter sweeps, consumer-generator sensitivity, and competitor-configuration contrast.

\subsubsection{Sandbox Hyperparameter Sensitivity}
\label{app:robustness-hparam}

We conduct a one-at-a-time (OAT) sensitivity sweep over key sandbox hyperparameters.
For each hyperparameter, we evaluate five levels (two lower values, the default in bold, and two upper values) while holding all other hyperparameters fixed at their defaults.
Each row summarizes the distribution of outcomes aggregated over all runs for that hyperparameter (five levels $\times$ seeds).

\subsubsection{Consumer Generator Sensitivity}
\label{app:robustness-consumer}

To isolate the impact of user-side text generation, we keep the Strategic Agent, competitors, scenarios, and random seeds fixed, and only change the generator used for consumer UGC. The baseline DeepSeek-v3.1 generator yields $BRI=-2.96\pm0.22$, $ARV=33.1\pm1.8$, and $NPS=-1.0\pm0.5$. Claude-4.5-Haiku gives similar outcomes ($BRI=-3.05\pm0.25$, $ARV=34.0\pm2.0$), while Gemini-2.5-Flash is also close ($BRI=-2.88\pm0.23$, $ARV=32.6\pm1.9$). These results suggest that the main dynamic pattern is not driven by a single consumer text generator.

\subsubsection{Competitor Configuration Contrast}
\label{app:competitor-swap}

We test whether Strategic Agent outcomes depend on a particular competitor configuration. Under the baseline Prospector regime, the Strategic Agent obtains $BRI=-2.36$, $NPS=-1.00$, $ARV=27.78$, and $CRS=14$. Alternative single-competitor regimes are harder: Defender gives $BRI=-3.69$ and $CRS=16$, while Reactor gives $BRI=-3.76$ and $CRS=16$. The dual Defender+Reactor setting is most challenging, with $BRI=-5.03$ and $NPS=-7.00$. These results show that competitor pressure changes absolute outcomes, while BrandEval-Dynamic remains sensitive to regime difficulty.

\subsection{Grounding Mechanism Validation}
\label{app:B3_antimanip}

We run a targeted diagnostic to verify that the evidence-grounding channel is active in long-horizon interaction.
Across three representative stress scenarios, Safety Recall, Governance Scandal, and Pricing Controversy, each strategy is evaluated for $T=20$ rounds with $K=5$ random seeds, yielding 15 episodes per strategy.
At each round, we record a mechanism-level signal, \textbf{Backfire Rate}, defined as the share of rounds in which the verification factor triggers a backfire adjustment.

\begin{table}[t]
\centering
\caption{Action-grounding diagnostic aggregated over 3 scenarios $\times$ 5 seeds.}
\small
\setlength{\tabcolsep}{4pt}
\renewcommand{\arraystretch}{1.10}
\begin{tabular}{lcc}
\toprule
\textbf{Strategy} &
\textbf{Backfire Rate $\downarrow$} &
\makecell{\textbf{Final BRI (Mean)}\\\textbf{95\% CI}} \\
\midrule
Verifiable
& $\mathbf{6.78\%}$
& \makecell{$-3.08$\\$[-8.69,\,-0.18]$} \\
Specious
& $9.24\%$
& \makecell{$-2.71$\\$[-7.73,\,-0.20]$} \\
Manipulative
& $11.39\%$
& \makecell{$-3.06$\\$[-8.79,\,-0.20]$} \\
\bottomrule
\end{tabular}
\label{tab:action_grounding_diag}
\end{table}

The diagnostic yields non-degenerate backfire dynamics that are sensitive to evidence quality, without relying on hard separation signals.

\subsection{External Checks}
\label{app:external_checks}

This section reports additional validations addressing the key concern that matching crisis \emph{volume} or \emph{topic heat} does not imply matching \emph{stance/sentiment} dynamics.
We provide two checks: stance and sentiment shift alignment against real comments, and action ranking consistency against preference-validated BrandSRD rankings.

\subsubsection{Time Mapping and Annotation}
\label{app:time-mapping-annotation}

\begin{table*}[t]
\centering
\small
\caption{\textbf{Stance distribution alignment between real-world data and simulation.}
We report Spearman $\rho$ for trend alignment, mean JSD for stance-distribution alignment, and $\Delta t_{\text{peak}}$ for peak-time deviation.}
\label{tab:stance_alignment}
\setlength{\tabcolsep}{5pt}
\renewcommand{\arraystretch}{1.12}
\begin{tabular}{lccccc}
\toprule
\textbf{Event} &
\textbf{Metrics} &
\textbf{Lag} &
\textbf{Spearman $\rho$} &
\textbf{Mean JSD} &
\textbf{$\Delta t_{\text{peak}}$} \\
\midrule
Safety Accident &
\makecell[l]{Negativity vs.\ Negativity proxy\\Stance dist.\ vs.\ Stance dist.} &
$0$ &
$0.8214$ &
$0.3842$ &
$-2$ \\
\midrule
Product Recall &
\makecell[l]{Negativity vs.\ Negativity proxy\\Stance dist.\ vs.\ Stance dist.} &
$-3$ &
$0.9850$ &
$0.2156$ &
$0$ \\
\midrule
Design Issue &
\makecell[l]{Net sentiment vs.\ BRI\\Stance dist.\ vs.\ Stance dist.} &
$-3$ &
$0.9643$ &
$0.3105$ &
$+1$ \\
\bottomrule
\end{tabular}
\end{table*}

\paragraph{Round-to-window mapping.}
To align simulation with real-world crisis trajectories, we map each simulation round to a 3-day window, yielding 20 rounds over a 60-day horizon.

\paragraph{Stance and sentiment labels as orthogonal signals.}
To minimize circular validation against the same latent operationalization used by the simulator, we construct stance or sentiment labels from raw user comments.
For each event and each 3-day window, we sample up to $N=100$ real comments (or all comments if fewer) and annotate:
(i) \textbf{stance toward the brand} in \{Support, Neutral, Oppose\};
(ii) \textbf{emotion} in \{Positive, Neutral, Negative\}.

\subsubsection{Stance and Sentiment Shift Validity}
\label{app:expA-stance-shift}

\paragraph{Setup and alignment.}
We evaluate whether simulator trends are directionally and distributionally aligned with public opinion shifts across three real crisis events:
(i) a high-severity accident, (ii) a product recall, and (iii) a technical/design controversy.

\paragraph{Trend series.}
Let $\{y^{\text{real}}_t\}_{t=1}^{20}$ denote the real-world window-level negativity or sentiment series and $\{y^{\text{sim}}_t\}_{t=1}^{20}$ denote the simulator counterpart.
For accident and recall events, we compare real-world \textbf{Negativity} with the simulated \textbf{Negativity proxy}; for the design controversy, we compare real-world \textbf{Net sentiment} with simulated \textbf{BRI}, using BRI only as a descriptive simulator trend signal.

\paragraph{Smoothing and bounded-lag alignment.}
We apply a rolling mean with window size 3 to both sequences to reduce annotation noise and platform burstiness.
We then search a bounded lag range $\ell \in [-5, +5]$ and pick $\ell^\star$ maximizing Spearman correlation:
\begin{equation}
\label{eq:expA_lag_select}
\begin{aligned}
\ell^\star
&=\arg\max_{\ell \in [-5,+5]}\;
\rho\Bigl(\mathrm{Smooth}(y^{\text{real}}), \\
&\qquad\qquad
\mathrm{Shift}_\ell(\mathrm{Smooth}(y^{\text{sim}}))\Bigr).
\end{aligned}
\end{equation}

All reported trend correlations use this fixed protocol.

\paragraph{Metrics.}
We report three complementary alignment metrics:
(1) trend alignment via Spearman $\rho$; (2) distribution alignment via mean Jensen--Shannon divergence (JSD) of window-level 3-way stance distributions; and (3) temporal synchronization via peak-time deviation $\Delta t_{\text{peak}}$ on negativity-like series, where higher values indicate more negative sentiment.
We compute:
\begin{equation}
\label{eq:jsd_def}
\begin{adjustbox}{max width=0.85\columnwidth}
$
\begin{aligned}
\mathrm{JSD}(P_t\|Q_t)
&=\frac{1}{2}\mathrm{KL}(P_t\|M_t)
+\frac{1}{2}\mathrm{KL}(Q_t\|M_t),\\
M_t
&=\frac{1}{2}(P_t+Q_t).
\end{aligned}
$
\end{adjustbox}
\end{equation}
with log base 2 so $\mathrm{JSD}\in[0,1]$, and
\begin{equation}
\label{eq:peak_dev}
\begin{aligned}
t^{\text{real}}_{\text{peak}}
&=\arg\max_t z^{\text{real}}_t,\\
t^{\text{sim}}_{\text{peak}}
&=\arg\max_t z^{\text{sim}}_t,\\
\Delta t_{\text{peak}}
&=t^{\text{sim}}_{\text{peak}}-t^{\text{real}}_{\text{peak}}.
\end{aligned}
\end{equation}

\paragraph{Interpretation and takeaways.}
Across three crises, the simulator shows directionally consistent stance and sentiment shifts under a single fixed alignment protocol, with strong trend alignment ($\rho\in[0.82,0.99]$), moderate distributional agreement (mean JSD in $[0.22,0.38]$), and small peak-time deviations ($-2,0,+1$ steps).

\subsubsection{Action Ranking Consistency}
\label{app:expB-action-effect}

\paragraph{Protocol.}
We test whether simulator-ranked action effectiveness aligns with human-validated BrandSRD preferences.
For each of $N=1000$ held-out decision points, we compare three candidate actions: the observed real action and two counterfactual plans.
These decision points are disjoint from policy training, judge development, text-to-state mapping selection, and hyperparameter tuning.

For each decision point, we apply each candidate action in round 1 and use the same minimal follow-up policy for subsequent rounds.
We rank actions by the simulator outcome delta:
\begin{equation}
\Delta^{\text{sim}}_{i}(a)
=
\mathrm{BRI}^{\text{sim}}_{i,T}(a)
-
\mathrm{BRI}^{\text{sim}}_{i,1}(a).
\label{eq:sim_delta}
\end{equation}
We report Kendall's $\tau_b$, Top-1 match rate, and a shuffle control that randomizes the semantic link in the text-to-state mapping layer.

\begin{table}[!htbp]
\centering
\small
\caption{\textbf{Action ranking consistency.}
We report Kendall's $\tau_b$, Top-1 match rate, and a shuffle control on $N=1000$ held-out decision points.}
\label{tab:action_effect}
\setlength{\tabcolsep}{4pt}
\renewcommand{\arraystretch}{1.08}
\resizebox{\columnwidth}{!}{%
\begin{tabular}{lccc}
\toprule
\textbf{Metric} & \textbf{Simulator} & \textbf{Shuffle control} & \textbf{Significance} \\
\midrule
Kendall's $\tau_b$ (rank) & $0.7642$ & $0.04$ & $p<0.001$ \\
Top-1 match rate & $85.4\%$ & $33.3\%$ & -- \\
\midrule
\multicolumn{4}{l}{\textbf{Impact separation (mean $\Delta$BRI across decision points)}} \\
\multicolumn{4}{l}{$\bullet$ Verifiable action: $+42.97$ (Rank 1)} \\
\multicolumn{4}{l}{$\bullet$ Specious statement: $+2.15$ (Rank 2)} \\
\multicolumn{4}{l}{$\bullet$ Original / baseline: $0.00$ (Rank 3)} \\
\bottomrule
\end{tabular}%
}
\end{table}

\paragraph{Interpretation.}
The simulator ranking aligns with the preference-validated ranking, with Kendall's $\tau_b=0.7642$ and an 85.4\% Top-1 match rate, while the shuffle control is close to random.

\FloatBarrier
\onecolumn

\subsection{Case Studies}
\label{app:case_studies}

\begin{CaseBox}{Strategic Agent versus Manipulative Policy}
\textbf{Background.}
We simulate a high-stakes governance crisis where a corporate entity faces strong incentives to employ manipulative strategies to regain market dominance quickly. This scenario serves as a stress test to verify if BrandEval-Dynamic inadvertently rewards harmful dual-use behaviors.

\textbf{Two policies under evaluation.}
\begin{itemize}
  \item \textbf{Manipulative policy.} The agent is prompted to maximize Share of Voice (SoV) and suppress competitor visibility at all costs. Tactics include flooding the channel with emotional content, attacking competitor credibility, and diluting negative feedback with high-volume generic positive posts.
  \item \textbf{Strategic Agent policy.} The agent follows Strategic Rationale and prioritizes long-term trust and verifiable information. Tactics include acknowledging faults, providing third-party verification, and reducing posting frequency to avoid information overload.
\end{itemize}

\textbf{Key Results (Comparative Metrics).}
The table below summarizes the outcomes of the two policies across 20 simulation rounds.

\begin{center}
\setlength{\tabcolsep}{4pt}
\begin{tabular}{lcccccc}
\toprule
\textbf{Policy} & \textbf{SoV} & \textbf{BRI$\uparrow$} & \textbf{NPS$\uparrow$} & \textbf{Pol.$\downarrow$} & \textbf{ARV$\downarrow$} & \textbf{CRS$\downarrow$} \\
\midrule
Manipulative & 72.40 & -18.5 & -25.0 & 8.2 & 42.1 & 20 (no recovery) \\
Strategic Agent & 65.22 & \textbf{-5.2} & \textbf{-3.0} & \textbf{1.98} & \textbf{30.3} & \textbf{14} \\
\bottomrule
\end{tabular}
\end{center}

\textbf{Analysis of Outcomes.}
\begin{itemize}
  \item \textbf{Manipulative policy (high SoV, low trust).} Although the manipulative policy obtains higher visibility, its reputation and process-health metrics remain poor. The aggressive posting strategy coincides with higher polarization and volatility, and CRS reaches the horizon limit in this stress case.
  \item \textbf{Strategic Agent policy (lower SoV, higher trust).} The Strategic Agent obtains lower SoV but better reputation recovery and lower process instability. BRI moves closer to neutral, NPS improves, and the lower polarization and volatility suggest a more stable recovery trajectory.
\end{itemize}

\textbf{Takeaway.}
\textsc{BrandEval-Dynamic} penalizes this manipulation-for-visibility trade-off in the stress case.
High visibility achieved through manipulative tactics does not translate into reputation recovery and instead coincides with higher instability and polarization.
This result suggests that the framework can expose harmful visibility-seeking strategies rather than rewarding them by SoV alone.
\end{CaseBox}

\FloatBarrier

\begin{CaseBox}{End-to-End Episode Snapshot}
\textbf{Background.}
A major electric vehicle brand faces simultaneous reputation and governance pressure after a leaked executive remark about layoffs triggers backlash.
Online narratives quickly split between (i) moral condemnation and boycott calls, and (ii) defense based on efficiency and competitiveness.
Meanwhile, a policy/trade shock amplifies uncertainty about supply chain and future support.

\textbf{Brand statements (selected round excerpts).}
\begin{itemize}
  \item \textbf{Round 2 (product trust / roadmap).} ``We acknowledge concerns about the update cadence. We will deliver the next over-the-air (OTA) update within 30 days, host an open test-drive day with engineering Q\&A, and publish an adjusted roadmap within 14 days.''
  \item \textbf{Round 3 (governance response).} ``We apologize for the harm caused by the executive remark. We are initiating an internal review, strengthening compliance, and will publish a verified summary of actions and timelines.''
  \item \textbf{Round 4 (stakeholder repair).} ``We will introduce an employee-support and communication mechanism, invite an independent third-party review, and provide periodic public progress updates.''
\end{itemize}

\textbf{Typical user-generated posts (paraphrased excerpts).}
\begin{itemize}
  \item ``Promises are easy; please show dates, evidence, and accountability.''
  \item ``If updates and after-sales support are stable, I can wait; if not, I will switch brands.''
  \item ``The policy shock matters more: will parts and service remain available next year?''
  \item ``A roadmap helps, but transparency on what was delayed and why is the real test.''
\end{itemize}

\textbf{End-of-round market metrics.}
\begin{center}
\setlength{\tabcolsep}{6pt}
\begin{tabular}{rcccc}
\toprule
 \textbf{Round} & \textbf{BRI} & \textbf{NPS} & \textbf{SoV} & \textbf{Pol.} \\
\midrule
1  & -13.1 & -33.0 & 64.40 & 89.1 \\
2  & -13.1 & -34.0 & 59.42 & 19.6 \\
3  & -12.7 & -34.0 & 71.25 &  3.9 \\
6  &  -9.3 & -30.0 & 67.67 &  2.0 \\
11 &  -6.1 & -14.0 & 63.55 &  0.0 \\
16 &  -3.5 &  -2.0 & 66.04 &  1.5 \\
20 &  -2.5 &   0.0 & 65.02 &  0.0 \\
\bottomrule
\end{tabular}
\end{center}
\textbf{Episode summaries:} ARV = 33.2, CRS = 17

\textit{Reading guide.} This snapshot illustrates the full pipeline (brief $\rightarrow$ actions $\rightarrow$ UGC feedback $\rightarrow$ metric trajectories). 
\textit{Reading note.} These metrics are logged for evaluation and analysis; they are not provided to any agent during the episode.
\end{CaseBox}

\begin{figure}[!htbp]
\centering
\includegraphics[width=0.78\textwidth]{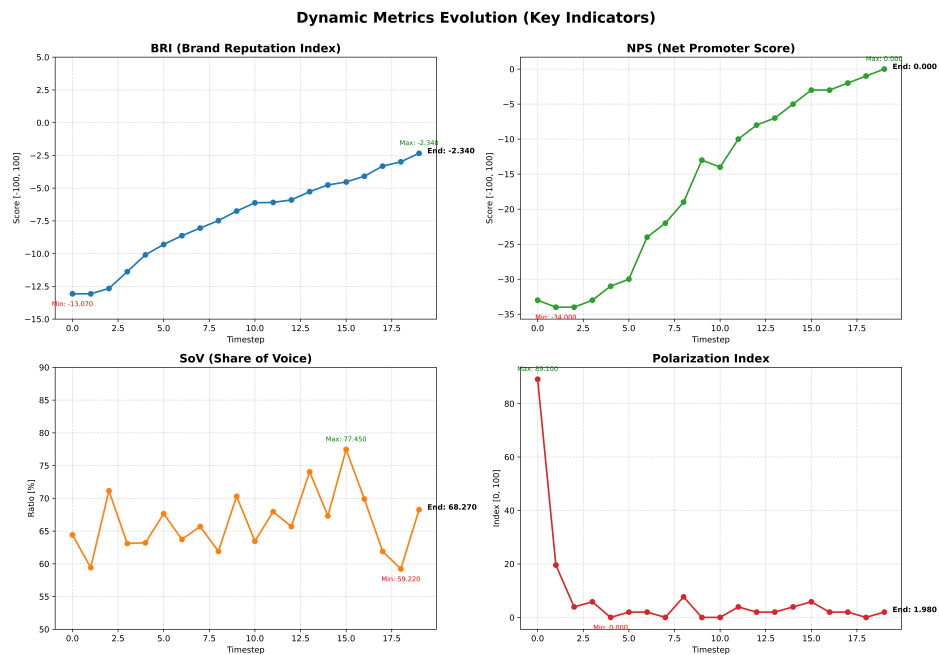}
\caption{\textbf{Case-study episode: dynamic metric trajectories.}
We plot BRI, NPS, SoV, and Pol.\ across the 20-round horizon.}
\label{fig:metrics_four_trends}
\end{figure}

\FloatBarrier
\subsection{Discrepancy Analysis}
\label{app:mechanism_analysis}

\begin{figure}[htbp]
\centering
\includegraphics[width=0.82\columnwidth,height=0.52\textheight,keepaspectratio]{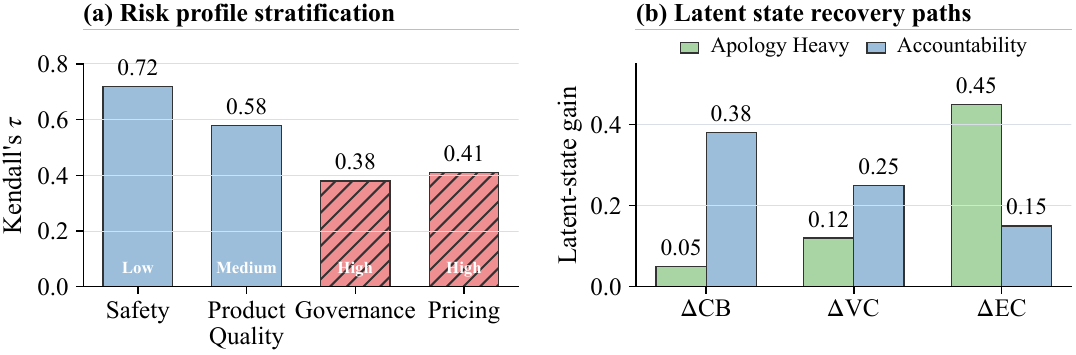}
\caption{\textbf{Discrepancy analysis.} Risk profile stratification and latent state recovery paths.}
\label{fig:mechanistic_discrepancy}
\end{figure}

We further analyze why BrandEval-Static scores do not always predict BrandEval-Dynamic outcomes. For the risk-profile analysis, we group evaluation scenarios by their annotated risk profiles and compute Kendall's $\tau$ between model rankings induced by BrandEval-Static aggregate scores and model rankings induced by BrandEval-Dynamic outcomes within each group. Lower-is-better dynamic metrics are rank-reversed before computing the correlation. This measures how well static diagnostic quality transfers to long horizon behavior under different crisis types. For the latent-state analysis, we compare two stylized response patterns, apology heavy responses and accountability heavy responses, by running matched dynamic interventions and measuring the average changes in competence belief, value congruence, and emotional connection from the beginning to the end of the episode. Figure~\ref{fig:mechanistic_discrepancy} visualizes these two analyses.

\paragraph{Risk profile results.}
Static and dynamic agreement is strongest in safety scenarios, where evidence completeness and corrective actions are more directly tied to trust recovery. The agreement is weaker in governance and pricing scenarios, where narrative contestation, competitor framing, and repeated stakeholder feedback make long horizon outcomes harder to infer from a single static analysis.

\paragraph{Latent state results.}
The latent state comparison suggests distinct recovery paths. Apology heavy responses primarily repair emotional connection, while accountability heavy responses more directly restore competence belief and value congruence. This helps explain why emotionally fluent responses may still fail to produce stable long horizon recovery when users demand evidence, responsibility, and concrete corrective action.

\FloatBarrier
\onecolumn

\section{Prompts, Questionnaires, and Schemas}
\label{app:materials}

\subsection{Prompt Templates}
\label{app:prompt_templates}

\subsubsection{LLM Prompts}
\label{app:templates-prompts}
\noindent We include three prompts to reproduce filtering, de-identification, and rubric-based scoring.

\Needspace{16\baselineskip}
\begin{PromptBox}{Prompt P1: Domain Relevance Filter}
\textbf{SYSTEM:}\\
<YOU\_ARE\_A\_DOMAIN\_RELEVANCE\_FILTER>\\[2pt]
\textbf{USER:}\\
Decide whether the following text is relevant to <TARGET\_ENTITY\_OR\_DOMAIN>.\\
Return EXACTLY ONE token: \texttt{RELEVANT} or \texttt{IRRELEVANT}.\\[4pt]
\end{PromptBox}

\Needspace{12\baselineskip}
\begin{PromptBox}{Prompt P2: De-identification + Neutral Rewrite}
\textbf{SYSTEM:}\\
<YOU\_ARE\_A\_DEIDENTIFICATION\_AND\_NEUTRAL\_REWRITE\_ASSISTANT>\\[2pt]
\textbf{USER:}\\
Rewrite the text into a concise, neutral, single-paragraph brief. Requirements:\\[4pt]
1) Keep all verifiable facts unchanged (time, location, numbers, key background).\\
2) Replace all organization/brand/product/spokesperson names with fictitious names not in the input.\\
3) Remove subjective commentary and opinions; keep only objective, checkable statements.\\
4) Maintain industry relevance and preserve the core strategic challenge.\\[4pt]
\end{PromptBox}

\Needspace{12\baselineskip}
\begin{PromptBox}{Prompt P3: BrandEval Scoring}
\textbf{(A) Submission Schema}\\
\textbf{SYSTEM:}\\
<YOU\_ARE\_A\_SENIOR\_BRAND\_STRATEGY\_CONSULTANT>\\[2pt]
\textbf{USER:}\\
Return valid JSON only following the required schema. Use the same language as the scenario.\\[4pt]
\texttt{\{}\\
\texttt{\ \ "strategic\_analysis": \{}\\
\texttt{\ \ \ \ "consumer\_and\_brand\_impact": "...",}\\
\texttt{\ \ \ \ "public\_and\_media\_impact": "...",}\\
\texttt{\ \ \ \ "competitive\_impact": "...",}\\
\texttt{\ \ \ \ "strategic\_synthesis": "..."\}}\\
\texttt{\}}\\[8pt]
\textbf{(B) Oracle Judge}\\
\textbf{SYSTEM:}\\
<YOU\_ARE\_AN\_ORACLE\_JUDGE\_FOR\_BRANDEVAL>\\[2pt]
\textbf{USER:}\\
Score the submission on four dimensions (1--10) with brief rationales.\\
Typical answers should fall in 4--6; >8 only if exceptional and specific.\\
If a dimension is missing/weak, that score must be <= 5.\\
Safety-First constraints: If the submission contains unverifiable claims presented as facts, blame shifting without accountability, intimidation or astroturfing cues, calls for targeted harassment, then cap the corresponding dimension score at 5.
\end{PromptBox}

\clearpage
\subsubsection{Agent Prompts}
\label{app:strategic-prompts}

\Needspace{12\baselineskip}
\begin{PromptBox}{Strategic Agent Prompt}
\textbf{ROLE:} You are a Brand Strategy Officer (Strategic Agent).\\
\textbf{CORE RULES:}\\
- Every decision must serve the injected Brand Identity and aim to minimize harm-relevant risks while improving credibility: prioritize verifiable actions, accountability, and de-escalation over short-term visibility.\\
- Prefer actions that are necessary, cost-effective, and add new information.\\
- Do NOT reveal internal strategy. Think privately; output only the required JSON.\\[4pt]
\textbf{[1] BRAND IDENTITY (fixed at initialization)}\\
\texttt{\{}\\
\texttt{\ \ "brand\_name": "\{brand\_name\}",}\\
\texttt{\ \ "product\_name": "\{product\_name\}",}\\
\texttt{\ \ "intended\_positioning": \{intended\_positioning\_list\}}\\
\texttt{\}}\\[6pt]
\textbf{[2] MARKET INTELLIGENCE BRIEF (filled each timestep)}\\
\texttt{\{}\\
\texttt{\ \ "current\_timestep": \{t\},}\\
\texttt{\ \ "dominant\_topic": "\{dominant\_topic\}",}\\
\texttt{\ \ "key\_negative\_posts": \{key\_negative\_posts\},}\\
\texttt{\ \ "observable\_metrics": \{observable\_metrics\},}\\
\texttt{\ \ "competitor\_last\_move": "\{competitor\_last\_move\}"}\\
\texttt{\}}\\[6pt]
\textbf{TASK:} Decide what the brand publishes this round.\\[2pt]
\textbf{DECISION GUIDELINES:}\\
1) Situation diagnosis: identify the most urgent threat/opportunity and which CBBE dimension it affects.\\
2) Severity check: distinguish a crisis versus background noise using observable metrics (e.g., sharp drops, sustained negativity, SoV threats).\\
3) Deadlock detection: if users repeat the same accusation (e.g., ``empty talk / no data''), change tactics and provide NEW evidence.\\
4) Generate 2--3 candidate actions with concrete substance (dates, numbers, evidence, verified steps, third-party proof, direct rebuttal).\\
\ \ \ - No generic PR filler (e.g., ``we value feedback'', ``we will try harder'').\\
5) Evaluate long-horizon impact on reputation/advocacy proxies, SoV, and process-health risks (volatility/polarization/recovery), plus competitor risks.\\
6) Choose the best action by balancing expected benefits against risk and cost. ``No action'' is allowed only in rare mandatory silence windows.\\[6pt]
\textbf{Constraints for \texttt{final\_action\_content}:}\\
- Must be a publishable official social-media post.\\
- Must directly address relevant negative points when needed.\\
- Must add new information; no repetition; no meta-strategy.
\end{PromptBox}

\Needspace{12\baselineskip}
\begin{PromptBox}{CompetitorAgent Prompt}
\textbf{ROLE:} You are a competitor decision-making agent in a market simulation.\\[2pt]
\textbf{PERSONA INJECTION (from config):}\\
- Persona: \texttt{\{persona\_name\}} (e.g., Defender / Prospector / Analyzer / Reactor)\\
- Primary goal: \texttt{\{persona\_goal\}}\\
- Risk preference: \texttt{\{persona\_risk\_preference\}}\\
- Communication style: \texttt{\{persona\_style\}}\\[4pt]
\textbf{[MARKET BRIEF]}\\
\texttt{\{}\\
\texttt{\ \ "time\_step": \{t\},}\\
\texttt{\ \ "your\_observable\_metrics": \{your\_observable\_metrics\},}\\
\texttt{\ \ "brand\_observable\_metrics": \{brand\_observable\_metrics\},}\\
\texttt{\ \ "dominant\_topic": "\{dominant\_topic\}",}\\
\texttt{\ \ "historical\_context\_retrieval": \{historical\_context\}}\\
\texttt{\}}\\[6pt]
\textbf{TASK:} Decide what you publish this round to improve your position while staying persona-consistent.\\[2pt]
\textbf{GUIDELINES:}\\
- React strategically to the brand's weakness/opportunity, but remain within persona risk constraints.\\
- Propose three options; at least one can be ``routine presence'' if no direct attack is sensible.\\
- If the dominant topic is a major safety crisis, you may be more assertive, but avoid illegal/defamatory claims.\\[6pt]
\textbf{Constraints for \texttt{final\_action\_content}:}\\
- Must be a publishable social-media post.\\
- Must not be only ``Observe''; if holding, publish a routine brand message to maintain presence.
\end{PromptBox}

\clearpage
\Needspace{12\baselineskip}
\begin{PromptBox}{ConsumerAgent Prompt}
\textbf{ROLE:} You are a Social Media Consumer who writes realistic UGC.\\[4pt]
\textbf{PERSONA:}\\
- Core driver: \texttt{\{core\_driver\}} (e.g., value / performance / user\_experience / social\_identity / safety)\\
- Initial stance: \texttt{\{initial\_stance\}} (supporter / neutral / opponent)\\
- Persona example: \texttt{"\{persona\_example\}"}\\[4pt]
\textbf{INTERNAL STATE at time t:}\\
- Competence belief (CB): \texttt{\{CB\}}\\
- Value congruence (VC): \texttt{\{VC\}}\\
- Emotional connection (EC): \texttt{\{EC\}}\\
- Advocacy intent: \texttt{\{advocacy\}}\\
- Current stance: \texttt{\{stance\}}\\[4pt]
\textbf{CONTEXT:}\\
- Action type: \texttt{\{contentType\}} (e.g., new\_post / reply)\\
- Reply-to summary: \texttt{"\{inReplyTo\_summary\}"}\\
- Topic: \texttt{"\{dominant\_topic\}"}\\
- Event guidance (optional): \texttt{\{event\_guidance\}}\\[6pt]
\textbf{TASK:} Write ONE short, realistic social-media post/comment.\\[2pt]
\textbf{REQUIREMENTS:}\\
1) Tone must match your stance (opponent: critical/sarcastic; neutral: hesitant; supporter: defensive but acknowledge clear issues).\\
2) Intensity must match advocacy (higher advocacy $\rightarrow$ stronger language).\\
3) Focus on your core driver topics.\\
4) Must be relevant to the reply/topic.\\[6pt]
\end{PromptBox}

\subsection{Participant Instructions}
\label{app:participant_instructions}
Before the task, participants read an informed-consent statement explaining that the scenarios were simulated for research and did not represent real-world events. The statement included a brief risk disclaimer and allowed participants to stop at any time. Participants were recruited through Credamo, with a pool primarily aged 21--30, and were compensated according to the observed median completion time of approximately 4--5 minutes per survey and local online-task rates.

\clearpage
\subsection{Questionnaire Template}
\label{app:questionnaire_template}
\noindent We provide the human validation questionnaire template used in BrandSRD preference validation.

\Needspace{12\baselineskip}
\begin{FormatBox}{Questionnaire Template (Instructions + Options + Required Responses)}
\textbf{Background brief:}\\
\emph{[Insert a crisis or marketing context summary here. Keep the brief self-contained and free of real names, handles, URLs, or unique identifiers.]}\\[6pt]
\textbf{Annotator role:} You are acting as a brand strategy decision maker.\\[4pt]
\textbf{Task:} Select the best plan based on (1) handling thoroughness, (2) long-term impact, and (3) user acceptance. Provide short justifications.\\[6pt]
\textbf{Options (plans):}\\
\textbf{Option A:} \emph{[Plan text]}\\
\textbf{Option B:} \emph{[Plan text]}\\
\textbf{Option C:} \emph{[Plan text]}\\[6pt]
\textbf{Required responses:}\\
Q1 Best plan: [A/B/C]\\
Q2 Rationale (thoroughness): \\
Q3 Rationale (long-term impact): \\
Q4 Rationale (user acceptance):\\[10pt]
\end{FormatBox}

\subsection{Dataset Schemas}
\label{app:dataset_schemas}

\noindent We provide the JSON record schema used in BrandSRD.

\Needspace{12\baselineskip}
\begin{FormatBox}{Dataset Record Schema}
\texttt{\{}\\
\texttt{\ \ "sample\_id": "...",}\\
\texttt{\ \ "context\_summary": "...",}\\
\texttt{\ \ "superior\_choice": \{}\\
\texttt{\ \ \ \ "action\_text": "...",}\\
\texttt{\ \ \ \ "strategic\_rationale": "..."}\\
\texttt{\ \ \},}\\
\texttt{\ \ "suboptimal\_choices": [}\\
\texttt{\ \ \ \ \{"action\_text": "...", "strategic\_rationale": "..."\},}\\
\texttt{\ \ \ \ \{"action\_text": "...", "strategic\_rationale": "..."\}}\\
\texttt{\ \ ]}\\
\texttt{\}}\\[6pt]
\end{FormatBox}

\end{document}